\documentclass[10pt,twocolumn,letterpaper]{article}

\usepackage{cvpr}              

\newcommand{\cmark}{\ding{51}}%
\newcommand{\thename}[0]{RAD-2}

\newcommand{\tablestyle}[2]{\setlength{\tabcolsep}{#1}\renewcommand{\arraystretch}{#2}\centering\footnotesize}
\usepackage{pifont}
\usepackage{multirow}
\usepackage[table]{xcolor}
\usepackage{makecell}
\usepackage{bbm}
\usepackage{marvosym} 
\usepackage{array}
\usepackage{makecell}
\usepackage{colortbl}
\usepackage{sectsty}
\usepackage{enumitem}
\usepackage{graphicx}

\usepackage{float} 
\usepackage{titlesec}

\titlespacing*{\section}
{0pt}
{10pt plus 2pt minus 2pt}  
{6pt plus 2pt minus 2pt}   

\titlespacing*{\subsection}
{0pt}
{8pt plus 2pt minus 2pt}
{4pt plus 2pt minus 2pt}

\definecolor{horizonblue}{RGB}{0, 102, 204}

\definecolor{cvprblue}{rgb}{0.21,0.49,0.74}
\definecolor{ourlightblue}{RGB}{245,247,255}
\usepackage[pagebackref,breaklinks,colorlinks,allcolors=horizonblue]{hyperref}
\def\thename{RAD-2}

\def\paperID{*****} 
\def\confName{CVPR}
\def\confYear{2026}

\definecolor{horizonblue}{RGB}{0, 102, 204}

\title{
    \vspace{-4em} 
    \noindent
    \makebox[\textwidth][s]{ 
        \includegraphics[height=2em]{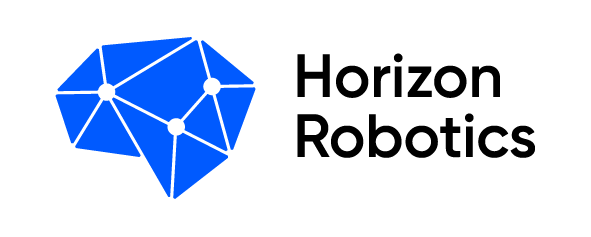} 
        \hspace{-0.5em}
        \raisebox{0.3em}{\color{gray!50}\rule{0.5pt}{1.2em}} \hspace{0.5em} 
        \includegraphics[height=2em]{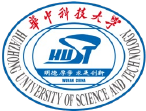} %
        \hfill 
    }
    \\
    \vspace{0.4em}
    {\color{horizonblue}\hrule height 1pt} 
    \vspace{1.0em}  
    \thename: Scaling Reinforcement Learning in a Generator-Discriminator Framework
    \vspace{1.0em} \\
    {\color{horizonblue}\hrule height 1pt} 

}

\author{
\textbf{Hao Gao}$^{1}$ \quad
\textbf{Shaoyu Chen}$^{2,\dagger}$ \quad
\textbf{Yifan Zhu}$^{2}$ \quad 
\textbf{Yuehao Song}$^{1}$ \\  
\textbf{Wenyu Liu}$^{1}$ \quad 
\textbf{Qian Zhang}$^{2}$ \quad 
\textbf{Xinggang Wang}$^{1,\textrm{\Letter}}$ \\
\textsuperscript{1}\,Huazhong University of Science \& Technology \quad
\textsuperscript{2}\,Horizon Robotics \\
\normalsize{Project page:
\url{https://hgao-cv.github.io/RAD-2}
}\\
}

\begin{document}
\maketitle

\let\thefootnote\relax\footnotetext{$^\dagger$ Project leader. $^\textrm{\Letter}$ Corresponding author.}

\begin{abstract}
High-level autonomous driving requires motion planners capable of modeling multimodal future uncertainties while remaining robust in closed-loop interactions. Although diffusion-based planners are effective at modeling complex trajectory distributions, they often suffer from stochastic instabilities and the lack of corrective negative feedback when trained purely with imitation learning. To address these issues, we propose \textbf{\thename}, a unified generator-discriminator framework for closed-loop planning. Specifically, a diffusion-based generator is used to produce diverse trajectory candidates, while an RL-optimized discriminator reranks these candidates according to their long-term driving quality. This decoupled design avoids directly applying sparse scalar rewards to the full high-dimensional trajectory space, thereby improving optimization stability. To further enhance reinforcement learning, we introduce Temporally Consistent Group Relative Policy Optimization, which exploits temporal coherence to alleviate the credit assignment problem. In addition, we propose On-policy Generator Optimization, which converts closed-loop feedback into structured longitudinal optimization signals and progressively shifts the generator toward high-reward trajectory manifolds. To support efficient large-scale training, we introduce BEV-Warp, a high-throughput simulation environment that performs closed-loop evaluation directly in Bird's-Eye View feature space via spatial warping. RAD-2 reduces the collision rate by 56\% compared with strong diffusion-based planners. Real-world deployment further demonstrates improved perceived safety and driving smoothness in complex urban traffic.
\end{abstract}
\vspace{-12pt}    
\begin{figure}[h]
\centering
\vspace{1mm}
\includegraphics[width=1.0\linewidth]{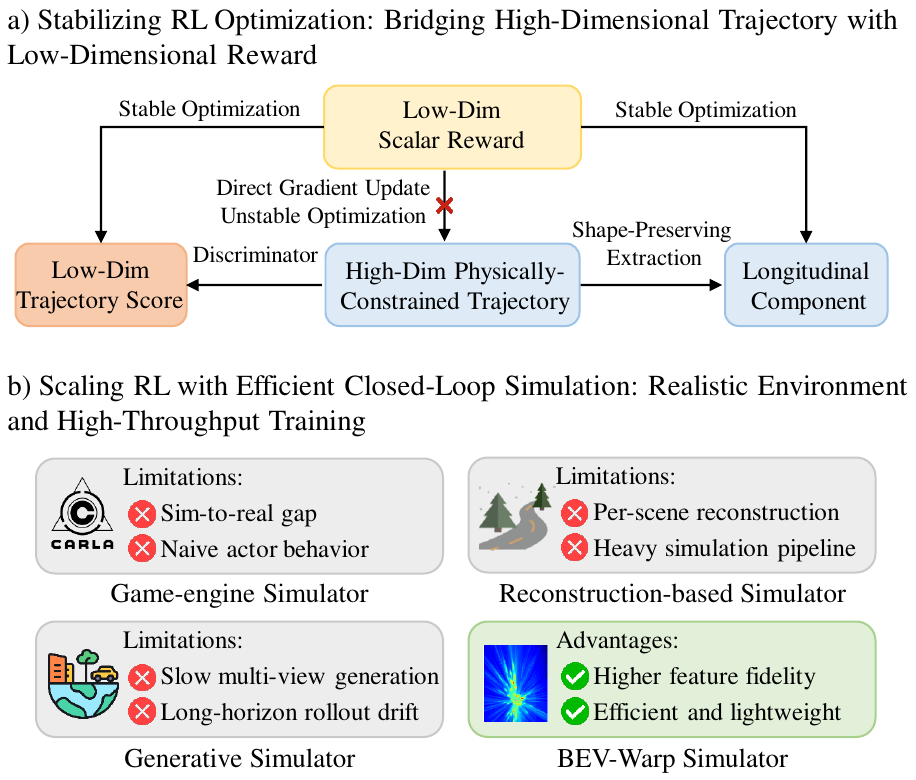} 
\caption{
\textbf{Scaling RL in a Generator-Discriminator framework.} (a) Stabilizing RL Optimization: RAD-2 projects the high-dim trajectory space into low-dim score and longitudinal components, ensuring stable policy updates. (b) Efficient Simulation: BEV-Warp enables high-throughput, feature-level closed-loop training, overcoming the limitations of existing simulators.}
\label{fig:rl-opti}
\end{figure}

\section{Introduction}
\label{sec:intro}
\begin{figure*}[ht]
\centering
\vspace{1mm}
\includegraphics[width=1.0\linewidth]{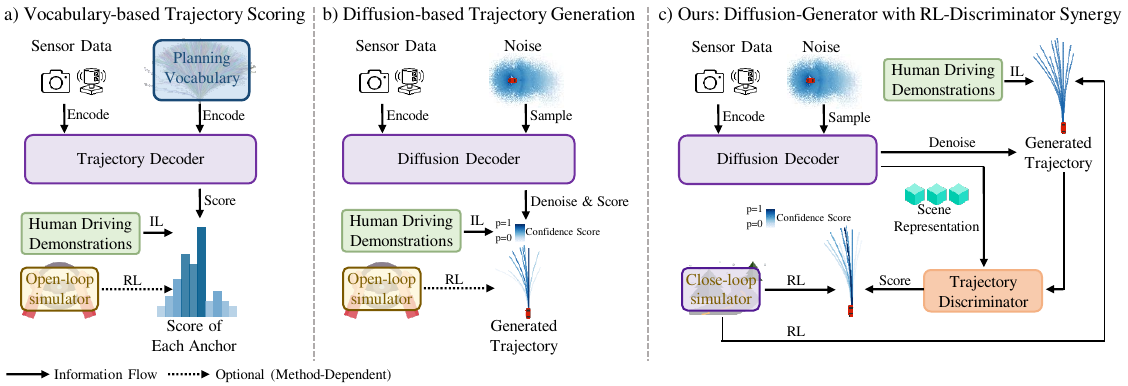} 
\caption{\textbf{The comparison of multimodal trajectory planning paradigms.} (a) Vocabulary-based scoring with fixed anchors~\cite{chen2024vadv2, li2024hydra}. (b) Diffusion-based multimodal generation~\cite{liao2025diffusiondrive}. (c) \thename{}: synergizing diffusion generator with RL-discriminator in closed-loop training.
}
\label{fig:teaser-2}
\end{figure*}

Achieving robust, safe, and human-like motion planning is a key goal in high-level autonomous driving systems. 
A variety of learning-based approaches have been explored for this purpose~\cite {kendall2019learning, chitta2022transfuser, wang2024driving}. 
Regression-based planners~\cite{jiang2023vad, hu2023uniad, weng2024paradrive} predict trajectories deterministically, thus collapsing multimodal behaviors and producing mean-biased outputs. 
Selection-based planners~\cite{chen2024vadv2, li2024hydra, li2025hydra, li2025ztrs} rely on a discrete set of candidates, limiting their ability to represent the full range of feasible trajectories.
To address these limitations, diffusion-based imitation learning (IL) planners~\cite{zheng2025diffusionplanner,liao2025diffusiondrive,zou2025diffusiondrivev2,zheng2025resad} have emerged as a promising approach in autonomous driving and embodied AI, providing a generative framework capable of modeling multimodal continuous trajectories and adapting them to the diverse conditions encountered in complex driving scenarios. 
An architectural comparison of different multimodal planning paradigms is provided in Fig.~\ref{fig:teaser-2}.

However, diffusion-based IL planners still face key challenges when applied to real-world autonomous driving. Real driving datasets contain noise and uneven distribution, which leads the diffusion model to learn certain regions of the trajectory distribution less effectively, resulting in occasional low-quality or unstable trajectories~\cite{lee2023refining, feng2024resisting}. This is especially critical for safety-sensitive planning. The high dimensionality of continuous trajectories further complicates optimization~\cite{li2025backtobasics, zheng2026unleashing}, and imitation-only training provides no negative feedback to suppress unrealistic behaviors.  In addition, IL introduces structural limitations: it suffers from causal confusion, learning correlations between states and actions instead of the underlying causal factors, which can lead to shortcut behaviors, and its open-loop training paradigm causes a mismatch with the closed-loop nature of real-world driving~\cite{gao2025rad, lu2023imitation}.

Several recent works have explored combining reinforcement learning (RL) with imitation learning (IL) to improve policy learning~\cite{gao2025rad,liu2025reinforced,li2025recogdrive,zhou2025autovla,liang2018cirl,lu2023imitation,huang2022efficient}. While RL provides task-level feedback beyond imitation, directly applying it to generative planners is challenging. The reward signal in RL is usually a low-dimensional scalar, whereas the action space involves high-dimensional, temporally structured trajectories, complicating credit assignment and making optimization unstable~\cite{lillicrap2015continuous}. RL training for autonomous driving requires closed-loop interaction, usually in simulation for safety and cost reasons. However, despite recent advances, existing simulation frameworks still exhibit several practical limitations. Game-engine simulators~\cite{fu2025minddrive,dosovitskiy2017carla} often simplify agent behavior, reconstruction-based simulators~\cite{kerbl20233dgs,gao2025rad,zhou2024hugsim,ni2025recondreamer-rl} are complex and costly, and learned world models~\cite{yan2025ad-r1, li2025omninwm, yang2026dreamerad, xia2025drivelaw, yang2026worldrft} struggle with long-horizon, multi-view generation. These limitations pose practical challenges for scaling RL to high-dimensional trajectory optimization.

To scale RL training, we propose a unified architecture that integrates a generator–discriminator design with a scalable simulation environment, termed BEV-Warp.
As illustrated in Fig.~\ref{fig:rl-opti}, this design enables stable policy updates and efficient closed-loop interaction.

Instead of directly optimizing the high-dimensional output of generator using sparse scalar rewards, which is challenging, we restrict reinforcement learning to the discriminator, whose output space naturally aligns with the low-dimensional reward signal.  Trained via closed-loop interaction, the discriminator learns to evaluate trajectories based on long-term outcomes, progressively improving its ability to distinguish high-quality trajectories from undesirable ones, thereby enhancing safety, efficiency, and comfort.

Improving system performance requires both accurate trajectory evaluation and refinement of the trajectory distribution. While the discriminator prioritizes higher-quality candidates, the generator determines the diversity and quality of behaviors that can be realized. To this end, we introduce On-policy Generator Optimization (OGO). Rather than applying reinforcement learning to the high-dimensional trajectory space, OGO constrains the optimization to the longitudinal component, where scalar rewards can optimize the policy without compromising stability.

The framework forms a self-improving closed loop. The discriminator $\mathcal{D}$ evaluates candidate trajectories, prioritizing higher-quality behaviors, while the generator $\mathcal{G}$ refines its trajectory distribution through on-policy optimization. By operating on the same interaction data, the two components achieve efficient joint optimization without additional simulation cost.
Formally, let $\tau$ denote a future trajectory. The diffusion generator $\mathcal{G}_\theta(\tau | o)$, parameterized by $\theta$, models a broad distribution of feasible actions conditioned on the context $o$. The RL-trained discriminator $\mathcal{D}_\phi(\tau | o, \mathcal{C})$, parameterized by $\phi$, provides a reranking distribution over a set of candidate trajectories $\mathcal{C} = \{\tau_1, \dots, \tau_M\}$ sampled from the generator, i.e., $\mathcal{C} \sim \mathcal{G}_\theta(\cdot | o)$. These two components define the joint policy distribution as
$\Pi_{\theta,\phi}(\tau | o) = \mathbb{E}_{\mathcal{C} \sim \mathcal{G}_\theta(\cdot | o)} [ \mathcal{D}_\phi(\tau | o, \mathcal{C}) ]$, aligning with the probabilistic inference framework for optimal control~\cite{levine2018reinforcement}.
This architecture inherently supports inference-time scaling~\cite{jaech2024openai}, where increasing the sample count $M$ allows the discriminator to explore a denser action space and identify higher-quality solutions without retraining. 
Through iterative optimization, the generator and discriminator jointly optimize the overall policy, progressively shifting the distribution toward safer and more efficient behaviors. This mutual improvement raises the system’s performance upper bound without relying on additional expert supervision.

In contrast to Large Language Models~\cite{brown2020language, guo2025deepseek}, autonomous driving involves high-dimensional continuous action spaces characterized by weak instantaneous reward-action correlations. This discrepancy leads to a severe credit assignment problem, as sparse scalar rewards fail to effectively distinguish which specific variations within a sampled group contribute to superior outcomes~\cite{li2025reinforcement}. To address this, we propose a Temporally Consistent (TC) sampling and optimization paradigm. During rollout collection, we employ a latched execution strategy, where a selected trajectory is reused over a fixed horizon to ensure behavioral coherence. Building upon this, we restructure the sampling process to enforce temporal dependencies across consecutive decision steps, ensuring that candidate trajectories are evaluated within a persistent behavioral context. This structured sampling allows us to introduce Temporally Consistent Group Relative Policy Optimization (TC-GRPO), which effectively denoises the advantage signals and stabilizes the policy gradient. Together, these mechanisms enable the generator–discriminator system to iteratively optimize trajectory quality and enhance closed-loop planning performance.

The contributions of \thename{} are summarized as follows:
\begin{itemize}
    \item We propose a unified generator-discriminator framework that stabilizes motion planning by decoupling diffusion-based trajectory exploration and RL-based reranking, ensuring robust performance in complex scenarios. 
    
    \item We introduce a joint policy optimization strategy that integrates TC-GRPO to ensure temporally consistent candidate reranking and OGO to iteratively shift the trajectory distribution via structured longitudinal optimizations.
    
    \item We develop an RL pipeline based on BEV-Warp, a high-throughput, feature-level simulation environment leveraging spatial equivariance for efficient policy iteration. \thename{} reduces the collision rate by over 56\% on large-scale benchmarks and significantly improves perceived safety during real-vehicle testing, yielding stable and comfortable planning behaviors.
\end{itemize}
\section{Related Work}
\subsection{Discriminator for Autonomous Driving}
Trajectory scoring and selection have emerged as pivotal techniques for enhancing the reliability of autonomous driving systems~\cite{liao2025diffusiondrive, guo2025ipad, zheng2025resad, zou2025diffusiondrivev2, sun2026sparsedrivev2}. Early works like VADv2~\cite{chen2024vadv2} and Hydra-MDP~\cite{li2024hydra, li2025hydra} rely on predefined trajectory vocabularies or rule-based teachers to guide selection. DriveSuprim~\cite{yao2025drivesuprim} further refines this paradigm with a coarse-to-fine filtering framework combined with self-distillation. Recent advances such as DriveDPO~\cite{shang2025drivedpo} and GTRS~\cite{li2025gtrs} incorporate preference optimization and dynamic candidate evaluation to improve flexibility. However, these discriminative approaches typically operate in an open-loop manner, often resulting in suboptimal decisions because they neglect long-term downstream consequences and are constrained by the diversity of discrete candidate sets. In contrast, \thename{} synergizes a continuous diffusion-based generator with a closed-loop trained discriminator, enabling robust planning over extended horizons by evaluating a more expressive manifold of future possibilities.

\subsection{RL for Autonomous Driving}
Reinforcement learning (RL)~\cite{schulman2017ppo,shao2024deepseekmath,yu2025dapo,zheng2025gspo} has been widely explored to mitigate the causal confusion and poor generalization issues of imitation learning. While recent works integrate RL with 3DGS-based digital twins~\cite{gao2025rad}, reasoning-oriented fine-tuning~\cite{jiang2025alphadrive}, or GRPO-based generation~\cite{li2025recogdrive, zhou2025autovla, lian2026fine}, optimizing high-dimensional driving outputs (e.g., raw trajectories) under sparse rewards remains notoriously difficult due to severe credit assignment challenges~\cite{liu2025reinforced, li2025recogdrive, zhou2025autovla, liang2018cirl, lu2023imitation, huang2022efficient}. Unlike these direct optimization approaches, we utilize RL rewards to train a low-dimensional trajectory discriminator, effectively reformulating the complex planning task into a tractable preference learning problem. This decoupling, empowered by TC-GRPO, leverages temporal coherence as a physical prior to stabilize the RL search space and ensure behavioral consistency. Finally, the generator is iteratively optimized via On-policy Generator Optimization to align with high-reward manifolds, achieving a superior balance between exploration and learning stability.

\subsection{Closed-loop Simulation Environment}
Closed-loop simulation is fundamental for the training and validation of RL-based driving policies. Traditional simulators like CARLA~\cite{dosovitskiy2017carla} and SMARTS~\cite{zhou2020smarts} offer interactive environments but often suffer from significant sim-to-real gaps due to their reliance on game engines. To bridge this gap, reconstruction-based simulators such as RAD~\cite{gao2025rad} and ReconDreamer-RL~\cite{ni2025recondreamer-rl} leverage 3D Gaussian Splatting (3DGS)~\cite{kerbl20233dgs} and video-diffusion priors to provide photorealistic training feedback. Furthermore, generative world models~\cite{hu2023gaia, wang2024driving-to-future, zhao2025drivedreamer4d, zhao2025recondreamer++, li2025wote} have been developed to synthesize future driving scenes or Bird's-Eye View (BEV) representations~\cite{philion2020lss, li2024bevformer} for trajectory evaluation. While generative approaches offer high fidelity, they are often computationally intensive and susceptible to cumulative temporal drift. To address these limitations, we introduce BEV-Warp, which enables high-throughput simulation by directly warping BEV features around the ego vehicle thereby bypassing the expensive image-level rendering process.
\section{Method}
\label{sec:method}
\subsection{Generator-Discriminator Framework} 
As illustrated in Fig.~\ref{fig:teaser-2} (c), our framework decomposes trajectory planning into two components: 
a diffusion-based generator that produces a diverse set of candidate trajectories, and a discriminator that evaluates and reranks these candidates. Together, they define a structured policy where generation and evaluation are jointly optimized.

\subsubsection{Diffusion-based Generator}  
The generator models a multimodal distribution over future trajectories and produces a set of candidate trajectories conditioned on the current observation \(o_t\), formally \(\mathcal{G}_\theta(\tau \mid o_t)\).

\noindent\textbf{Scene Encoding.} The observation \(o_t\) is first encoded into BEV features \(T_{b}\), capturing the spatial layout of the scene. Static scene elements (e.g., lane geometry, road boundaries, and other map structures) and dynamic agents (e.g., surrounding vehicles and pedestrians with their motion states) are extracted from this representation to provide a comprehensive understanding of the environment.  Let \(\mathcal{X}_{\text{map}}\), \(\mathcal{X}_{\text{agent}}\), and \(\mathcal{X}_{\text{nav}}\) denote the predicted static map elements, dynamic agents, and provided navigation inputs (e.g., waypoints or reference path), respectively. Token embeddings are then obtained via lightweight encoders:
\begin{equation}
T_{m} = \mathrm{E}_{m}(\mathcal{X}_{\text{map}}), ~
T_{a} = \mathrm{E}_{a}(\mathcal{X}_{\text{agent}}), ~ 
T_{n} = \mathrm{E}_{n}(\mathcal{X}_{\text{nav}}).
\end{equation}

These embeddings are fused with \(T_b\) through a learnable module \(\mathcal{F}(\cdot)\) to obtain the unified scene embedding:
\begin{equation}
E_{\text{scene}} = \mathcal{F}(T_b, T_m, T_a, T_n),
\end{equation}
which conditions the DiT-based trajectory generator via cross-attention.

\noindent\textbf{Trajectory Generation.}
For \(M\) independent modes, an initial noise trajectory \(\tau^{(0,m)} \sim \mathcal{N}(0,I)\) is iteratively denoised for \(K\) steps:
\begin{equation}
\tau^{(k,m)} = G\!\big(\tau^{(k-1,m)}, E_{\text{scene}}, k\big), \quad k = 1, \dots, K,
\end{equation}
where \(G\) denotes the conditional denoising network~\cite{ho2020denoising} that predicts the next trajectory sample given the current noisy sample \(\tau^{(k-1,m)}\), the scene embedding \(E_{\text{scene}}\), and the current step \(k\). The final set of candidate trajectories is
\begin{equation}
\widehat{\mathcal{T}} = \big\{\hat{\tau}^{m}_{t:t+H}\big\}_{m=1}^M, 
\quad \hat{\tau}^{m}_{t:t+H} = \tau^{(K,m)} \sim \mathcal{G}_\theta(\tau \mid o_t),
\end{equation}
where \(\hat{\tau}^{m}_{t:t+H}\) is a continuous \((x,y)\) trajectory over the planning horizon \(H\). These trajectories define the candidate set of the policy and are passed to the discriminator for evaluation and selection.

\subsubsection{RL-based Discriminator}
\label{subsubset:scorer}
The discriminator evaluates candidate trajectories based on their expected outcomes under the current scene, providing a preference over the candidate set.

\noindent\textbf{Trajectory Encoding.}
Each point in the trajectory is embedded via a shared MLP, yielding a sequence of embeddings $\{e_i\}_{i=1}^H$.  
The trajectory sequence, prepended with a learnable \textsc{[cls]} token, is processed by a Transformer encoder. The \textsc{[cls]} output aggregates information from the entire trajectory and serves as the trajectory-level query \(Q_{\tau}\).

\noindent\textbf{Scene Conditioning.}
The discriminator constructs a scene representation from the same inputs \(\mathcal{X}_{\text{map}}\) and \(\mathcal{X}_{\text{agent}}\) as the generator.  
Its static and dynamic encoders, \(\mathrm{E}^{*}_{m}\) and \(\mathrm{E}^{*}_{a}\), share the same architecture as the corresponding generator encoders \(\mathrm{E}_{m}\) and \(\mathrm{E}_{a}\), but maintain independent parameters:

\begin{equation}
T^{*}_{m} = \mathrm{E}^{*}_{m}(\mathcal{X}_{\text{map}}), ~
T^{*}_{a} = \mathrm{E}^{*}_{a}(\mathcal{X}_{\text{agent}}).
\end{equation}

\noindent\textbf{Trajectory–Scene Interaction.}
The trajectory query \(Q_{\tau}\) aggregates multi-source scene context via cross-attention $\Psi(Q, KV)$:
\begin{equation}
\begin{aligned}
O_{\mathrm{m}} = &\Psi(Q_\tau, T^*_m),
O_{\mathrm{b}} = \Psi(Q_\tau, T_b),
O_{\mathrm{a}} = \Psi(Q_\tau, T^*_a), \\
&T_{a \cap m} = \Psi(T^*_a, T^*_m),
O_{\mathrm{a \cap m}} = \Psi(Q_\tau, T_{a \cap m}).
\end{aligned}
\end{equation}
These embeddings are aggregated as $E_{\text{fusion}}$:
\begin{equation}
E_{\text{fusion}} = \Phi([O_{\mathrm{m}}; O_{\mathrm{b}}; O_{\mathrm{a}}; O_{\mathrm{a \cap m}}]), 
\end{equation}
where $\Phi(\cdot)$ denotes concatenation followed by an MLP.

\noindent\textbf{Trajectory Scoring and Reranking.}
For each candidate trajectory \(\hat{\tau}_{t:t+H} \in \widehat{\mathcal{T}}\), the discriminator produces a scalar score via the sigmoid activation $\sigma$ applied to the fused representation:
\begin{equation}
s(\hat{\tau}_{t:t+H}) = \sigma(E_{\text{fusion}}) \in [0,1].
\end{equation}
Optionally, the scores can be normalized across the candidate set to induce a distribution over trajectories for reranking. In our implementation, we directly use the sigmoid outputs for prioritization.

\subsection{Closed-Loop Simulation Environment and Controller}
The proposed policy is evaluated in an efficient BEV-Warp environment, which enables high-throughput interaction by transforming BEV features. To verify the generalization across varied scene representations, validation is also conducted in a 3D Gaussian Splatting (3DGS) environment.

\noindent\textbf{BEV-Warp Environment.}
BEV-Warp constructs a simulation environment by directly manipulating BEV features over time, bypassing redundant image-level rendering. The simulation is initialized from recorded real-world sequences, including multi-view observations, ego-vehicle trajectories, and scene context.

For each simulation step $t$, the system extracts the BEV feature $\mathcal{B}^{\text{ref}}_{t}$ and loads the recorded pose $\mathcal{P}_{t} \in \mathrm{SE}(2)$ as the state. The planner generates candidate trajectories $\widehat{\mathcal{T}}$, from which an optimal trajectory $\hat{\tau}^*$ is selected and tracked to update the agent's simulated pose $\mathcal{P}_{t+1}$.

\begin{figure}[t]
\centering
\vspace{1mm}
\includegraphics[width=1.0\linewidth]{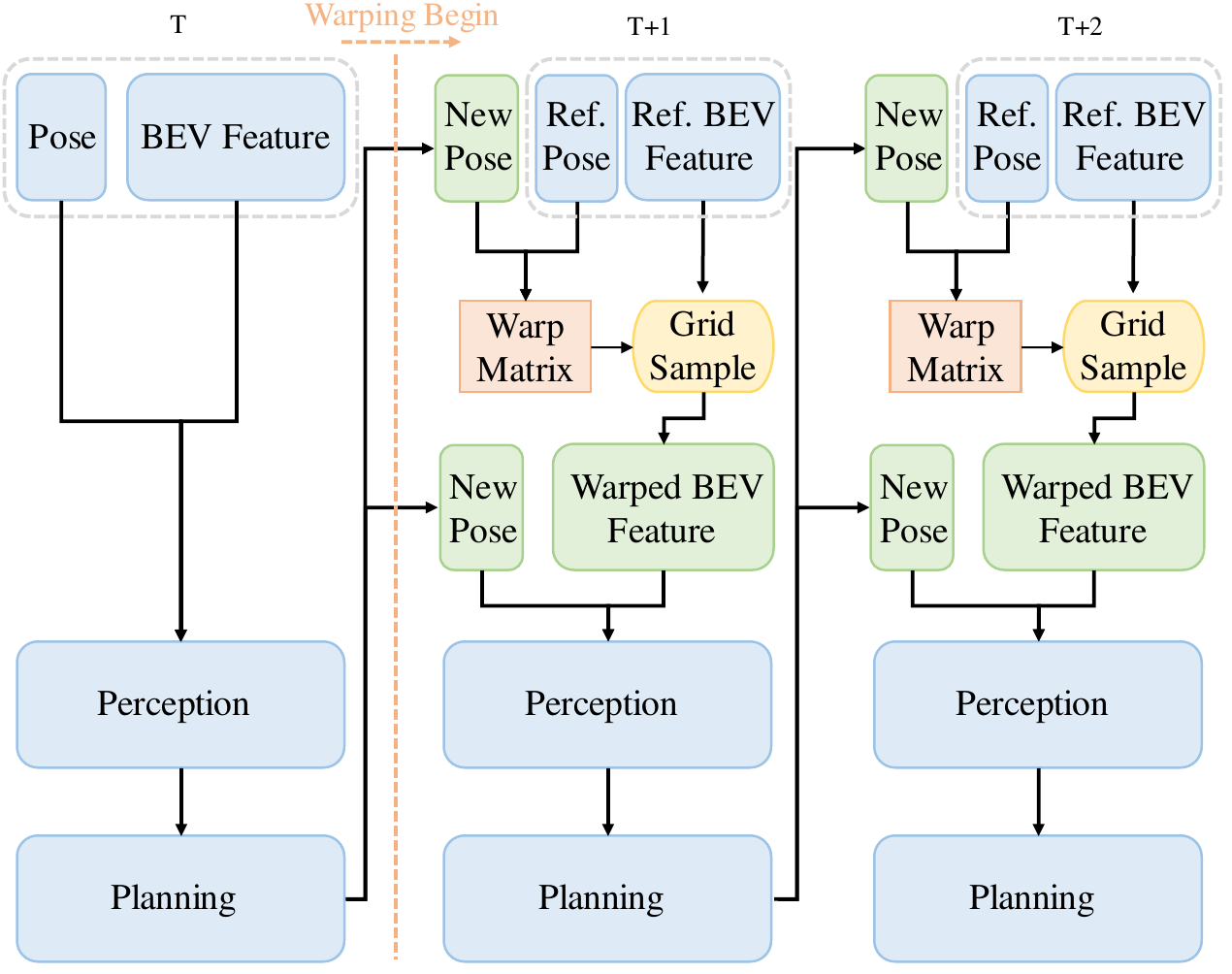} 
\caption{\textbf{Schematic of the BEV-Warp simulation environment.} 
The closed-loop evaluation is driven by a recursive feature-warping mechanism. 
At each timestep, a transformation matrix $\mathbf{M}_{t}$ is derived from the relative pose deviation between the simulated agent $\mathcal{P}_{t}$ and the logged reference $\mathcal{P}^{\text{ref}}_{t}$. 
This matrix is then applied to the reference BEV feature $\mathcal{B}^{\text{ref}}_{t}$ via spatial warping, synthesizing a high-fidelity observation $\mathcal{B}_{t}$ for the subsequent planning cycle without expensive image-level rendering.}
\label{fig:bevwarp}
\end{figure}

As illustrated in \textbf{Fig.~\ref{fig:bevwarp}}, to realign the logged environment with the simulated ego-state, we derive a warp matrix $\mathbf{M}_{t+1} = (\mathcal{P}_{t+1})^{-1} \mathcal{P}^{\text{ref}}_{t+1} \in \mathbb{R}^{3 \times 3}$. 
Assuming constant altitude and neglecting rotations (i.e., pitch and roll),
the synthesized feature for the next timestep is obtained via:
\begin{equation}
\mathcal{B}_{t+1} = \mathcal{W}\left(\mathcal{B}^{\text{ref}}_{t+1}, \mathbf{M}_{t+1}\right),
\end{equation}
where $\mathcal{W}(\cdot)$ is implemented via bilinear interpolation \cite{jaderberg2015spatial}.
This synthesized representation, coupled with $\mathcal{P}_{t+1}$, enables a continuous interaction for closed-loop evaluation.

\noindent\textbf{3DGS Environment.} 
To ensure performance consistency across different scene representations, the policy is further validated in a photorealistic simulation world built via 3D Gaussian Splatting (3DGS). Unlike the feature-level warping, this setup renders raw multi-view observations $o_t$ directly from the scene $G$ using the agent's current pose $\mathcal{P}_t$: $o_t = \mathcal{R}(G, \mathcal{P}_t)$, where $\mathcal{R}(\cdot)$ is the differential rendering operator. This serves as a testbed for verifying the policy's robustness against explicit image-based rendering.

\noindent\textbf{Trajectory-Based Controller.} 
Both environments employ an iLQR-based controller~\cite{li2004iterative} to track the planned trajectory $\hat{\tau}^*$. Given the non-linear vehicle dynamics $\mathbf{x}_{k+1} = f(\mathbf{x}_k, \mathbf{u}_k)$ (e.g., a kinematic bicycle model), the controller minimizes a quadratic cost function over the horizon $H$:
\begin{equation}
    \mathbf{u}^*_{t:t+H} = \arg\min_{\mathbf{u}} \sum_{k=t}^{t+H} \left( \|\mathbf{x}_k - \hat{\mathbf{x}}_k^{\text{ref}}\|_Q^2 + \|\mathbf{u}_k\|_R^2 \right),
\label{eq:ilqr}
\end{equation}
where $\hat{\mathbf{x}}_k^{\text{ref}}$ is the reference state derived from $\hat{\tau}^*$. The iLQR algorithm iteratively refines the control sequence via local linearization of $f(\cdot)$ and quadratic approximation of the cost. Specifically, the first optimal command $\mathbf{u}_t^*$ is applied to update the vehicle state, yielding the new ego pose $\mathcal{P}_{t+1}$ for the subsequent simulation step.

\subsection{Joint Policy Optimization}
\begin{figure*}[h]
\centering
\includegraphics[width=1.0\textwidth]{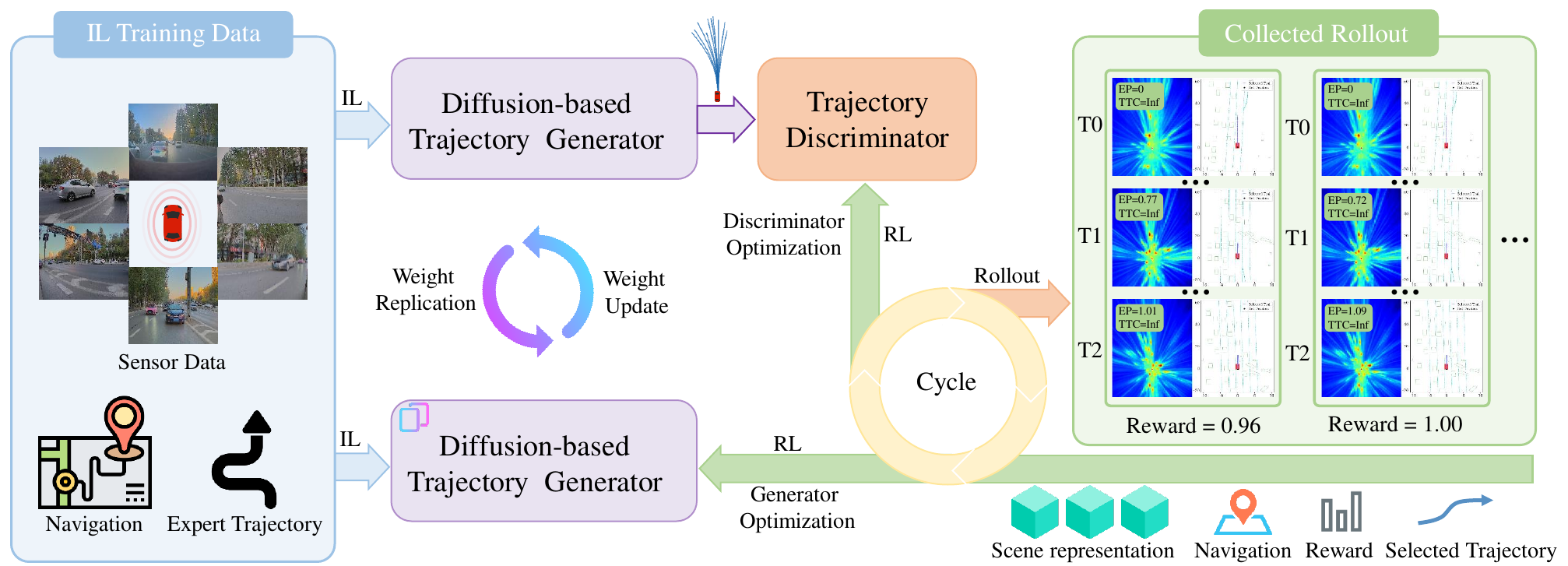} 
\caption{\textbf{The RAD-2 training pipeline.} Our approach synergizes a Diffusion-based generator $\mathcal{G}$ and a Transformer-based discriminator $\mathcal{D}$ within a multi-stage optimization loop: 
      (a) \textbf{Pre-training Stage:} $\mathcal{G}$ is initialized via imitation learning to capture multi-modal trajectory priors from expert demonstrations. 
      (b) \textbf{Closed-loop Rollout:} The joint policy, integrating $\mathcal{G}$ for generation and $\mathcal{D}$ for selection, interacts with the high-throughput BEV-Warp environment to generate diverse rollout data.
      (c) \textbf{Discriminator Optimization:} $\mathcal{D}$ is optimized via Temporally Consistent Group Relative Policy Optimization, leveraging closed-loop feedback to enhance its scoring precision. 
      (d) \textbf{Generator Optimization:} Through On-policy Generator Optimization, $\mathcal{G}$ is optimized via structured longitudinal optimizations derived from low-reward rollouts, effectively shifting its distribution toward safer and more efficient driving behaviors.}
\label{fig:framework}
\vspace{-10pt}
\end{figure*}

To optimize the policy within the BEV-Warp environment, we propose a joint optimization framework that synergizes the diffusion generator $\mathcal{G}_\theta$ with an RL-trained discriminator $\mathcal{D}_\phi$ through a cyclic structured pipeline. The global objective is to minimize the KL-divergence between the hybrid policy $\Pi_{\theta,\phi}$ and the ideal risk-neutral, high-efficiency distribution $\Pi^*$:
\begin{equation}
    \min_{\theta, \phi} \mathbb{D}_{\text{KL}} \big( \Pi_{\theta,\phi}(\tau \mid o) \parallel \Pi^*(\tau \mid o) \big).
\end{equation}
As illustrated in Fig.~\ref{fig:framework}, this objective is realized via a three-stage iterative process: 
(i) Temporally Consistent Rollout, which collects stable closed-loop interaction data to ground exploratory behaviors (Sec.~\ref{subsec:traj_reuse}); 
(ii) Discriminator Optimization, where $\phi$ is optimized via trajectory-level RL to internalize sparse rewards and provide precise ranking feedback (Sec.~\ref{subsec:rl_optimization}) ; and 
(iii) Generator Optimization, which employs On-policy Generator Optimization to concentrate the density of $\mathcal{G}_\theta$ onto high-reward manifolds (Sec.~\ref{subsec:ogo}).

\subsubsection{Temporally Consistent Rollout}
\label{subsec:traj_reuse}
While the planning head generates multimodal trajectory hypotheses at each timestep, re-sampling and switching between diverse modes at high frequency often disrupts the continuity of the agent's motion intentions. In the context of reinforcement learning, such frequent mode-switching decuples the correlation between a specific policy decision and its long-term driving outcome, leading to inefficient credit assignment and unstable policy improvement~\cite{zhang2022generative}.

To mitigate these issues, we implement a trajectory reuse mechanism to maintain short-term behavioral consistency. Once an optimal trajectory $\hat{\tau}^*_t$ is selected at time $t$, it is converted into a structured control sequence $\mathbf{u}_{t:t+H}^*$ following the iLQR optimization defined in Eq.~\eqref{eq:ilqr}. Instead of re-planning at $t+1$, the policy reuses the pre-computed controls over a fixed execution horizon $H_{\text{reuse}} < H$. Specifically, for each relative offset $h \in \{0, \dots, H_{\text{reuse}}-1\}$, the corresponding control command $\mathbf{u}_{t+h}^*$ is executed sequentially, ensuring that the vehicle state evolves consistently along the committed trajectory.
While $H_{\text{reuse}}$ is set as a fixed hyperparameter to ensure sampling efficiency, this mechanism natively supports asynchronous termination if safety constraints are violated, ensuring that the trajectory reuse does not compromise the agent's reactive capabilities in dynamic environments. By stabilizing the exploratory direction, this mechanism ensures that the cumulative reward more accurately reflects the quality of the selected trajectory $\hat{\tau}^*_t$, thereby facilitating effective policy gradients.

\subsubsection{Discriminator Optimization via RL}
\label{subsec:rl_optimization}
To guide the discriminator in capturing long-horizon driving semantics, we establish a multi-objective reward function to evaluate closed-loop outcomes and optimize the policy via a Temporally Consistent Group Relative Policy Optimization (TC-GRPO) framework. 

\noindent\textbf{Reward Modeling.} 
The reinforcement learning environment is built upon a collection of long-duration clips. Within each clip, the policy $\Pi_{\theta,\phi}$ interacts with the environment in a closed-loop manner to generate multiple rollouts. Each rollout represents an integrated decision-making chain, for which we define two complementary rewards to characterize safety and efficiency:

(i) \textit{Safety-Criticality Reward ($r_{\text{coll}}$):} We quantify collision risks via counterfactual interpolation. Specifically, at each simulation step $t$, the Time-To-Collision $T_t$ is defined as the earliest moment of intersection between the ego-vehicle's projected occupancy and the environment's ground-truth state:
\begin{equation}
T_t = \inf \{ k \in [0, T_{\text{max}}] \mid \mathcal{B}_{\text{ego}}(k; t) \cap \mathcal{V}_{\text{env}}(t+k) \neq \emptyset \},
\end{equation}
where $T_{\text{max}}$ is the safety detection threshold, $\mathcal{B}_{\text{ego}}(k; t)$ denotes the ego-vehicle's occupancy at look-ahead $k$ (projected from step $t$), and $\mathcal{V}_{\text{env}}(t+k)$ represents the collective ground-truth occupancy of the environment at absolute time $t+k$. If no intersection is detected, $T_t$ defaults to $T_{\text{max}}$. The sequence-level reward is then formulated as the worst-case temporal margin:
\begin{equation}
r_{\text{coll}} = \min_{1 \le t \le L} \left( T_t - T_{\text{max}} \right),
\end{equation}
where $L$ is the rollout horizon. This bottleneck-style formulation ensures that any momentary safety violation within the rollout dominates the sequence-level reward, thereby enforcing a strictly conservative driving policy.

(ii) \textit{Navigational Efficiency Reward ($r_{\text{eff}}$):} To synchronize the traversal pacing with expert demonstrations, we evaluate the sequence-level \textit{Ego Progress} (EP), denoted as $\rho$, at the conclusion of each rollout. The reward $r_{\text{eff}}$ is formulated as a stability window penalty to anchor the vehicle's progress within a target efficiency interval $[\rho_{\text{low}}, \rho_{\text{high}}]$:
\begin{equation}
r_{\text{eff}} = \min(\rho - \rho_{\text{low}}, 0) + \min(\rho_{\text{high}} - \rho, 0) + 1,
\end{equation}
where $\rho_{\text{low}} = 1.05$ and $\rho_{\text{high}} = 1.10$ define the normalized progress boundaries relative to the clipped expert trajectory. This formulation treats navigational efficiency as a bounded synchronization task, penalizing both sluggish under-performance and overly-aggressive deviations to enforce a human-like traversal pace.

Together, they balance safety and driving efficiency in closed-loop driving.

\noindent\textbf{Optimization Objective.} To address the credit assignment problem in continuous driving space, we propose Temporally Consistent Group Relative Policy Optimization (TC-GRPO). This framework introduces a structured rollout and reward assignment mechanism specifically designed for the temporal dependencies of motion planning, where sampled trajectory hypotheses are persisted over a short horizon to ensure behavioral coherence. By ensuring that the sparse environment reward is directly attributed to the specific trajectory hypothesis sustained within each persistent interval, our approach prevents the reinforcement signal from being diluted by high-frequency switching between disparate motion intentions. This temporal alignment effectively denoises the policy gradient and stabilizes the iterative optimization of the generator-discriminator system.

For a rollout $\mathcal{O}_i$ with sequence-level reward $r_i$, the standardized advantage $A_{i}$ is computed relative to the group $\{\mathcal{O}_i\}_{i=1}^G$ generated from the same initial state:
\begin{equation}
A_{i} = \frac{r_{i} - \text{mean}(\{r_{1}, \dots, r_{G}\})}{\text{std}(\{r_{1}, \dots, r_{G}\})}.
\end{equation}

We define $\mathcal{K}_{i}$ as the set of timesteps in rollout $\mathcal{O}_i$ where a new trajectory $\hat{\tau}^*_{i,t}$ is sampled to commence a latch-execution interval. The clipped objective for these sparse decision points is formulated as:
\begin{equation}
\mathcal{L}_{i,t \in \mathcal{K}_{i}} = \min\Big(\rho_{i,t} A_{i},\; \text{clip}(\rho_{i,t}, 1-\varepsilon, 1+\varepsilon) A_{i}\Big),
\end{equation}
where $\rho_{i,t} = \frac{\mathcal{D}_\phi(\hat{\tau}^{*}_{i,t} \mid o_{i,t})}{\mathcal{D}_{\phi_{\text{old}}}(\hat{\tau}^{*}_{i,t} \mid o_{i,t})}$ is the importance sampling ratio. This formulation ensures that the advantage signal reinforces the coherent trajectory hypothesis rather than independent action samples that lack temporal coherence.

To mitigate premature convergence and sigmoid saturation, we introduce an adaptive entropy regularization mechanism with temperature-based control. Concretely, let $\mathcal{H}_{i,t}$ denote the policy entropy at timestep $t$ of the $i$-th rollout, and $\bar{\mathcal{H}}$ denote its batch-wise average. The adaptive regularization weight is computed based on the average entropy:
\begin{equation}
    \beta = \exp(\lambda) \cdot \mathbbm{1}_{[\bar{\mathcal{H}} < \bar{\mathcal{H}}_{\text{target}}]},
\end{equation}
where $\lambda$ is a learnable temperature parameter and $\bar{\mathcal{H}}_{\text{target}}$ denotes the target entropy level. This formulation activates the entropy regularization exclusively when the average entropy falls below the target. The temperature parameter $\lambda$ is adaptively tuned via gradient descent only under such deficit conditions.

The RL objective for the discriminator is formulated as:
\begin{equation}
    \mathcal{J}_{\text{RL}}(\phi) = \mathbb{E} \Bigg[ \frac{1}{\sum_{i=1}^G |\mathcal{K}_i|} \sum_{i=1}^G \sum_{t=1}^{|\mathcal{K}_i|} \big(\mathcal{L}_{i,t} + \beta \mathcal{H}_{i,t}\big) \Bigg].
\end{equation}

By integrating adaptive entropy regularization with $\mathcal{L}_{i,t}$, the discriminator ensures stable exploration without compromising policy stability.

\subsubsection{On-policy Generator Optimization}
\label{subsec:ogo}
The hybrid policy $\Pi_{\theta,\phi}$ is defined by the composition of generator $\mathcal{G}_\theta$ and discriminator $\mathcal{D}_\phi$. While Sec.~\ref{subsec:rl_optimization} optimizes $\mathcal{D}_\phi$ for candidate selection, the mode coverage of $\mathcal{G}_\theta$ constrains the explorable policy space. To expand this boundary, we optimize $\mathcal{G}_\theta$ via On-policy Generator Optimization, which transforms closed-loop feedback into structured longitudinal optimization signal and repositions the generator's probability mass toward regions with favorable long-term outcomes.

\noindent\textbf{Reward-Guided Longitudinal Optimization.} 
At each closed-loop time step $t$, we obtain a raw lookahead trajectory segment $\tau^{\text{raw}}_t = \{(x_{t+h}, y_{t+h}, v_{t+h}, a_{t+h})\}_{h=0}^{H}$ over a horizon $H$. Based on the closed-loop feedback, we identify potential collision risks or insufficient progress and optimize the longitudinal component accordingly. This process adjusts the acceleration profile by applying a constant offset to the original values (with sign-aware handling) and re-integrating. This yields an optimized trajectory $\tau^{\text{opt}}_t$ that preserves the spatial path of $\tau^{\text{raw}}_t$ while optimizing its temporal progression to better align with the reward signals.

(i) \textit{Safety-driven Deceleration:} When the Time-to-Collision $T_t$ at the current time step falls below a threshold $\gamma_{\text{safe}}$, we apply temporal compression by reducing the travel distance over the horizon $H$ by a fixed ratio $\rho \in (0,1)$.

(ii) \textit{Efficiency-driven Acceleration:} 
When the ego vehicle lags behind the reference trajectory (indicating insufficient progress) and the instantaneous Time-to-Collision $T_t$ exceeds a safety threshold (i.e., no collision risk), we apply temporal extension by increasing the travel distance over the horizon $H$ by a fixed ratio $\rho' > 1$.

By converting closed-loop reward signals into structured optimizations, this mechanism enables the generator to iteratively shift its output distribution toward safer and more efficient behaviors.

\noindent\textbf{On-policy Distribution Shifting.} 
The optimized trajectory segments $\tau^{\text{opt}}_t$ are aggregated into an on-policy dataset $D_{\text{opt}} = \{\tau^{\text{opt}}_t\}$, which provides structured supervision for fine-tuning the generator $\mathcal{G}_\theta$ via a mean squared error loss over the prediction horizon $H$:
\begin{equation}
\mathcal{L}_{\text{op}}(\theta) = \mathbb{E}_{\tau^{\text{opt}} \sim D_{\text{opt}}} \left[ \sum_{k=0}^{H} \big\| \hat{\tau}_{t+k} - \tau^{\text{opt}}_{t+k} \big\|_2^2 \right].
\end{equation}
This distribution shift is inherently gradual and stable, as the target trajectories in $D_{\text{opt}}$ are constructed through on-policy interaction and dimension-specific optimization.
\section{Experiment}
\subsection{Dataset Details}
\noindent\textbf{Generator Pretraining.}
For diffusion generator pretraining, we leverage approximately $ 50,000$ hours of real-world driving data that provide ego-vehicle trajectories across diverse traffic scenarios. These trajectories are used to pretrain the motion generator, enabling the model to capture the multimodal distribution of human driving behaviors before downstream optimization.

\noindent\textbf{Closed-loop Training and Evaluation in the BEV Warp Environment.} We first collect 50k clips from real-world logs, each representing a continuous driving segment of 10–20 seconds. These clips cover diverse driving conditions and are categorized based on high-level driving objectives, including safety-oriented scenarios (e.g., interactions with elevated collision risk) and efficiency-oriented scenarios (e.g., driving under varying traffic conditions with suboptimal efficiency). The clips are evaluated in closed-loop simulation within the BEV Warp environment, where the pre-trained multimodal trajectory generator produces multiple candidate trajectories at every simulation step, from which one is randomly sampled to advance the ego vehicle. Based on the resulting outcomes, clips are filtered to retain safety-critical scenarios with elevated collision risk and efficiency-related scenarios with suboptimal driving performance. From these filtered scenarios, we curate two separate training sets corresponding to safety-oriented and efficiency-oriented objectives, each containing 10k clips, which are used for closed-loop RL training. For evaluation, we similarly construct two disjoint subsets of challenging clips corresponding to safety-oriented and efficiency-oriented scenarios, respectively, each containing 512 clips, for closed-loop assessment.

\noindent\textbf{Closed-loop Training and Evaluation in the 3DGS Environment.}
In addition to the BEV Warp environment, we further train and evaluate the trajectory discriminator in the photorealistic 3D Gaussian Splatting (3DGS) simulation benchmark introduced in Senna-2~\cite{song2026senna2}, which focuses on safety-oriented driving scenarios involving high-risk interactions. Among them, 1,044 clips are used for training and 256 clips are reserved for closed-loop evaluation.

\noindent\textbf{Open-loop Evaluation Scenarios.}
We also adopt the open-loop evaluation dataset from Senna-2~\cite{song2026senna2}. The benchmark covers six representative driving scenarios: car-following start, car-following stop, lane changing, intersections, curves, and heavy braking. These scenarios span diverse driving conditions, including longitudinal control, lane-level interactions, and complex road geometries, enabling a comprehensive evaluation of planning quality across different driving contexts.

\subsection{Experiment Setup}
\noindent\textbf{Baseline.} We adopt the same perception backbone as RAD~\cite{gao2025rad}, which encodes multi-view images into BEV features and extracts static scene and dynamic agent tokens, providing inputs to the planning modules. The planning head follows the modeling formulation of ResAD~\cite{zheng2025resad}, which is based on a standard diffusion framework for generative trajectory prediction. This planning head is used as the baseline trajectory generator in our experiments.

We compare our method with several representative planning approaches, including waypoint regression planners~\cite{chitta2022transfuser,jiang2023vad,zheng2024genad}, diffusion-based generative planners~\cite{zheng2025resad}, and selection-based planners~\cite{chen2024vadv2}. All methods are implemented and trained on our dataset using the same perception backbone, differing only in the planning head used to model future trajectories.

\begin{figure}[t]
\centering
\vspace{1mm}
\includegraphics[width=1.0\linewidth]{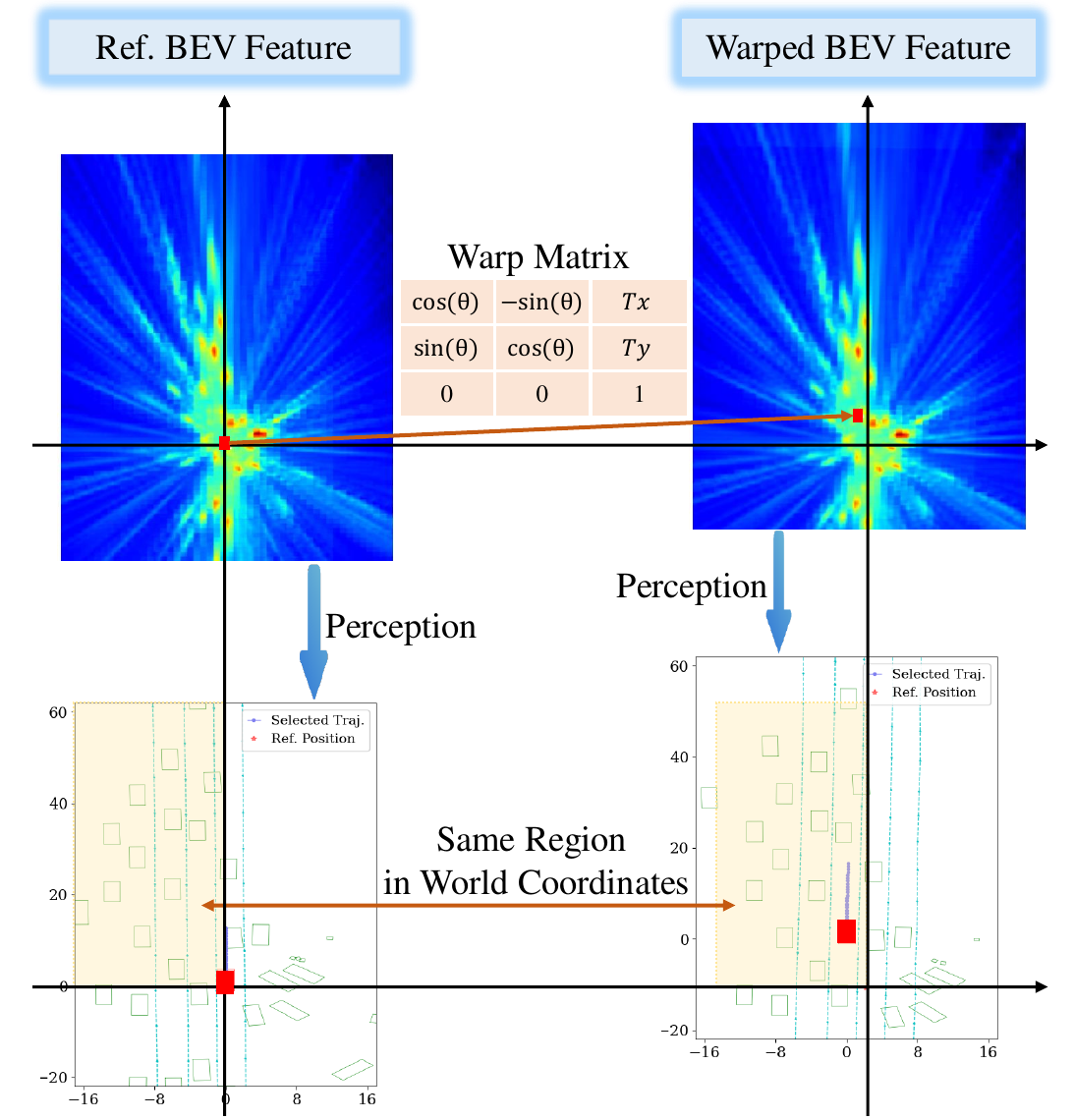} 
\caption{\textbf{Spatial equivariance in BEV-Warp.} We evaluate the reliability of warped features via a frozen pre-trained perception head. \textbf{(Left)} Reference feature $\mathcal{B}^{\text{ref}}$ and its corresponding perception outputs. \textbf{(Right)} Under ego-displacement (e.g., $\theta = 1.08^{\circ}, T_x = 10.658\text{m}, T_y = 2.387\text{m}$), the warped features yield perception results that maintain precise spatial alignment. This consistency across significant spatial transformations confirms that BEV-Warp preserves complex semantic geometry (e.g., dynamic agents and lane topologies) requisite for high-fidelity closed-loop simulation.}
\label{fig:warp_correctness}
\end{figure}

\noindent\textbf{Training Stages.} RAD-2 follows a multi-stage training:

(i) \textit{Pre-training:} Following RAD~\cite{gao2025rad}, the model first performs perception and planning pre-training. In the perception stage, the backbone encodes multi-view images into BEV features and extracts static scene and dynamic agent tokens, forming structured scene representations for downstream planning. 
Crucially, these BEV features exhibit strong spatial equivariance, ensuring that geometric transformations in the feature space correspond strictly to physical movements in the real world. As demonstrated in Fig.~\ref{fig:warp_correctness}, applying a warp $\mathbf{M}$ to the reference feature $\mathcal{B}^{\text{ref}}$ results in a predictable shift of the decoded perception outputs (e.g., 3D bounding boxes) that remains consistent with the simulated ego-motion while maintaining stability in the global coordinate system. 
In the planning stage, these representations are fed into a diffusion-based planning head trained on expert demonstrations to establish a baseline for human-aligned trajectory generation.

(ii) \textit{Reinforcement Learning:} After pre-training, the model is further optimized in simulation by collecting rollouts using the current policy. The collected trajectories are stored in a shared buffer for coordinated training of both components: the discriminator is trained via reinforcement learning to assign higher scores to safe and efficient trajectories, while the generator is iteratively optimized using structured longitudinal signals derived from the collected rollouts. By leveraging this shared self-generated data, the system progressively enhances trajectory quality, safety, and closed-loop performance.

\begin{figure}[t]
\centering
\vspace{1mm}
\includegraphics[width=1.0\linewidth]{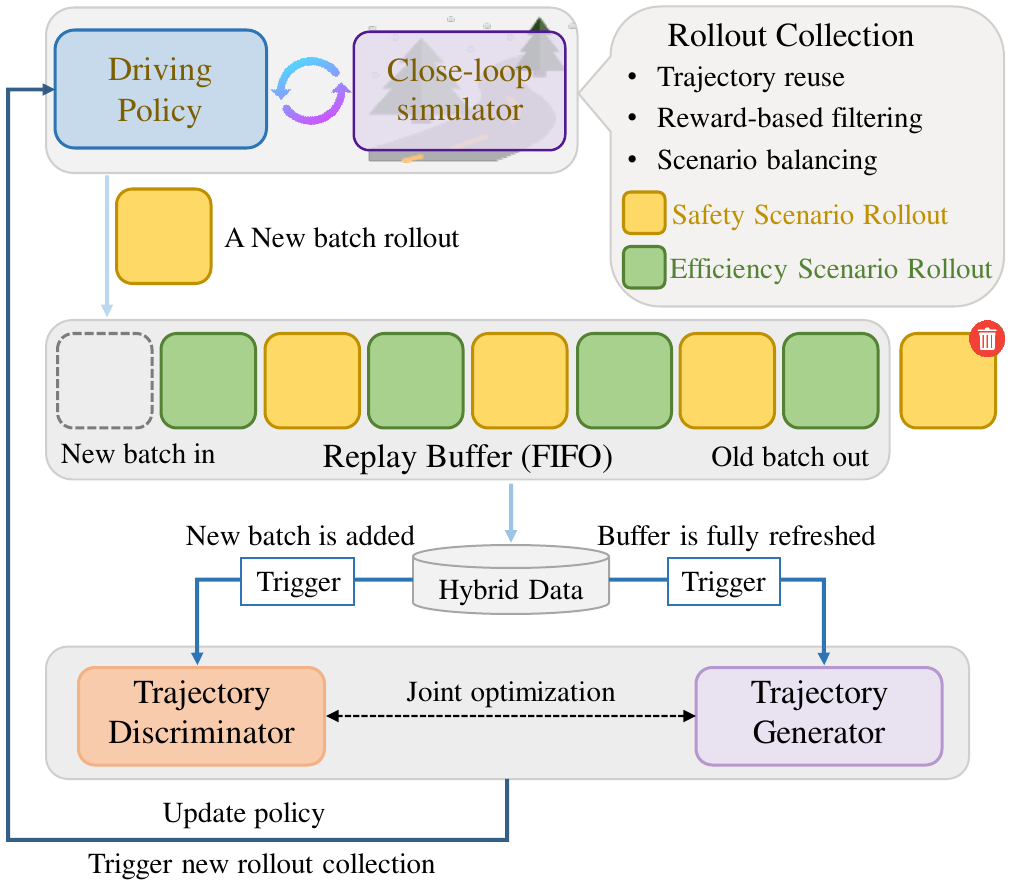} 
\caption{\textbf{Replay buffer management and closed-loop optimization workflow.} 
Driving rollouts are archived in a FIFO replay buffer that maintains a curated balance between safety-critical (yellow) and efficiency-oriented (green) scenarios. 
The system employs an asynchronous dual-trigger optimization strategy: 
(i) the trajectory discriminator is updated upon each batch ingestion to maintain sensitivity to recent rollout quality; 
(ii) the trajectory generator undergoes optimization only after a full buffer circulation (every 8 batches).}
\label{fig:databuffer}
\end{figure}

\noindent\textbf{Reinforcement Learning Implementation Details.} During the reinforcement learning stage, multiple driving rollouts are collected under the current policy for each clip. Trajectory reuse is employed to preserve short-term behavioral consistency (as described in Sec.~\ref{subsec:traj_reuse}), ensuring stable exploration during training. For each clip, we compute the mean and standard deviation of rollout rewards and discard the entire clip if the standard deviation falls below a predefined threshold. This filtering ensures that only clips with sufficiently diverse trajectory outcomes contribute to reinforcement learning. All valid rollouts are stored in a fixed-length replay buffer of 8, maintained in a first-in-first-out (FIFO) manner. Fig.~\ref{fig:databuffer} provides a comprehensive visualization of the data flow and the balanced data composition within the FIFO buffer.

During training, batches of rollouts are sampled from the buffer in groups of 4, and the standardized advantage is computed across each group.

The discriminator model is initialized by loading weights from the planning head for components that share the same architecture, such as the static and dynamic encoders, as detailed in Sec.~\ref{subsubset:scorer}. Trajectory-leve policy gradients are applied to optimize the discriminator, with an entropy regularization term included to prevent the collapse of trajectory scores to extreme values.

The optimization alternates between the discriminator and the generator in a closed-loop fashion. Each time a new batch of rollouts is added, the discriminator is optimized using the current buffer contents. Once the buffer has been fully refreshed with 8 batches of new rollouts, a generator optimization is performed, resulting in an approximate 8:1 training frequency between the discriminator and generator. This ensures continuous co-adaptation of the discriminator and generator while maintaining diverse training data.

\begin{table*}[ht]
    \setlength{\tabcolsep}{8.5pt}\centering
    \caption{\textbf{Closed-loop performance comparison in the BEV-Warp simulation environment.} Our proposed approach achieves the lowest collision rates and highest navigation efficiency, demonstrating superior closed-loop interactive capabilities.}
    \resizebox{\linewidth}{!}{
    \begin{tabular}{l|cccccccc} 
        \toprule
        \multirow{2}{*}{Method} 
        & \multicolumn{4}{c}{\textbf{Safety-oriented Scenario}} 
        & \multicolumn{3}{c}{\textbf{Efficiency-oriented Scenario}} \\
        \cmidrule(lr){2-5} 
        \cmidrule(lr){6-8} 
        & CR $\downarrow$ 
        & AF-CR $\downarrow$ 
        & Safety@1 $\uparrow$ 
        & Safety@2 $\uparrow$ 
        & EP-Mean $\uparrow$
        & EP@1.0 $\uparrow$
        & EP@0.9 $\uparrow$ \\
        \midrule
        TransFuser~\cite{chitta2022transfuser} & 0.563 & 0.275 & 0.400 &\underline{0.346} & 0.897 & 0.244 & 0.531 \\
        VAD~\cite{jiang2023vad}               &0.594 &0.299 &0.371 &0.312 &0.904 &0.252 &0.623 \\
        GenAD~\cite{zheng2024genad}           &0.592 &0.305 &0.363 &0.309 &0.930 &0.467 &0.736 \\
        ResAD~\cite{zheng2025resad}           &\underline{0.533} &\underline{0.264} & \underline{0.418} &0.281 & \underline{0.970} & \underline{0.516} & \underline{0.894} \\
        \rowcolor{ourlightblue}
        \textbf{\thename{}}                   &\textbf{0.234} &\textbf{0.092} &\textbf{0.730} &\textbf{0.596} & \textbf{0.988} &\textbf{0.736} & 0.984 \\
        \bottomrule
    \end{tabular}
    }
    \label{tab:bevwarp-benchmark}
\end{table*}

\begin{table}[h]
    \centering
    \tablestyle{6.6pt}{1.0}
    \caption{\textbf{Evaluation on photorealistic 3DGS benchmark.} Our method outperforms IL and RL baselines in safety-critical scenarios. Notably, it achieves superior risk mitigation and safety margins, as reflected by the lower Collision Rate (CR) and higher Safety@1/2 scores.}
    \resizebox{\linewidth}{!}{
    \begin{tabular}{l|cccccccccccc}
        \toprule
        Method                       & CR $\downarrow$ & AF-CR $\downarrow$ & Safety@1 $\uparrow$ & Safety@2 $\uparrow$ \\
        \midrule
        TransFuser~\cite{chitta2022transfuser} & 0.435           & 0.269              & 0.531               & 0.454               \\
        VAD~\cite{jiang2023vad}               & 0.502           & 0.280              & 0.458               & 0.362               \\
        VADv2~\cite{chen2024vadv2}               & 0.422           & 0.199              & 0.514               & 0.458               \\
        GenAD~\cite{zheng2024genad}           & 0.557           & 0.244              & 0.402               & 0.332               \\
        ResAD~\cite{zheng2025resad}           & 0.509           & 0.288              & 0.469               & 0.399               \\
        Senna~\cite{jiang2024senna}           & 0.310           & 0.111              & 0.638               & 0.539               \\
        Senna-2~\cite{song2026senna2}           & \underline{0.269}  & \textbf{0.077}     & \underline{0.667}      & \underline{0.565}               \\
        RAD~\cite{gao2025rad}           & 0.281  & 0.113     & 0.613      & 0.543               \\
        \rowcolor{ourlightblue}
        \textbf{\thename{}}                & \textbf{0.250}  & \underline{0.078}     & \textbf{0.723}      & \textbf{0.644}      \\
        \bottomrule
    \end{tabular}
    }
    \label{tab:3dgs-benchmark}
\end{table}

\noindent\textbf{Metrics.} 
We evaluate the closed-loop performance of our framework across two fundamental dimensions, safety and efficiency, using metrics derived directly from the simulated outcomes.
To assess collision avoidance and risk management, we report the Collision Rate (CR) and the At-Fault Collision Rate (AF-CR), where the latter specifically accounts for incidents attributable to ego-vehicle decision errors. 
The robustness of safety margins is further quantified by Safety@1s and Safety@2s, representing the proportion of clips where the minimum Time-to-Collision (TTC) remains above 1 and 2 seconds, respectively. 
Concurrently, navigation effectiveness is evaluated through the Ego Progress Mean (EP-Mean), which measures the average ratio of distance traveled relative to the reference route. 
We also include EP@1.0 and EP@0.9 as indicators of task completion reliability, denoting the percentage of scenarios where the vehicle successfully fulfills 100\% and 90\% of its assigned navigation goals, respectively.

For open-loop evaluation, we adopt the metrics from Senna-2~\cite{song2026senna2}, including Final Displacement Error (FDE) and Average Displacement Error (ADE) for trajectory accuracy, along with Collision Rate (CR), Dynamic Collision Rate (DCR), and Static Collision Rate (SCR) for safety assessment of predicted trajectories against dynamic and static obstacles.

\begin{table}[t]
    \centering
    \tablestyle{1.6pt}{1.0}
    \caption{\textbf{Open-loop benchmark of Senna-2.} Our approach achieves state-of-the-art trajectory accuracy with the lowest ADE and FDE. The consistently minimal collision risk across diverse scenarios further validates the precision of our generative distribution in aligning with expert driving priors.}
    \resizebox{\linewidth}{!}{
    \begin{tabular}{l|ccccc}
        \toprule
        Method 
        & FDE (m) $\downarrow$ 
        & ADE (m) $\downarrow$ 
        & CR (\%) $\downarrow$ 
        & DCR (\%) $\downarrow$ 
        & SCR (\%) $\downarrow$  \\
        \midrule
        TransFuser~\cite{chitta2022transfuser} 
        & 0.844 & 0.297 &0.981 & 0.827 & 0.154  \\
        VAD~\cite{jiang2023vad}               
        & 0.722 & 0.262 &0.621 & 0.554 & 0.067  \\
        GenAD~\cite{zheng2024genad}           
        & 0.806 & 0.290 &0.520 & 0.491 & 0.030 \\
        ResAD~\cite{zheng2025resad}           
        & 0.634 & 0.234 &0.378 & 0.367 & 0.011 \\
        Senna~\cite{jiang2024senna}           
        & 0.633 & 0.236 &0.294 & 0.286 & 0.008  \\
        Senna-2~\cite{song2026senna2} & \underline{0.597} & \underline{0.225} &\underline{0.288} & \underline{0.283} & \underline{0.005}\\
        \rowcolor{ourlightblue}
        \textbf{\thename{}} & \textbf{0.553} & \textbf{0.208} & \textbf{0.142} & \textbf{0.138} & \textbf{0.004} \\
        \bottomrule
    \end{tabular}
    }
    \label{tab:ol-comparison}
\end{table}

\begin{table*}[ht]
    \centering
    \tablestyle{6.6pt}{1.0}
    \setlength{\tabcolsep}{4.0pt}
    \caption{\textbf{Ablation study of the training pipeline.} We evaluate the contribution of Imitation Learning (IL), On-policy Generator Optimization, and Discriminator Optimization. The results demonstrate that while IL provides essential priors, the synergistic optimization of the generator and discriminator is critical for balancing the safety-efficiency trade-off.}
    \label{tab:ablation_training_stages}
    \begin{tabular}{c| cccc|ccccccc}
        \toprule        
        \multirow{2}{*}{\textbf{ID}} & \textbf{Gen. Pre-training} & \multicolumn{2}{c}{\textbf{Gen. Fine-tuning}} & \textbf{Disc. Training} & \multicolumn{4}{c}{\textbf{Safety-oriented Scenario}} & \multicolumn{3}{c}{\textbf{Efficiency-oriented Scenario}} \\
        \cmidrule(lr){2-2} \cmidrule(lr){3-4} \cmidrule(lr){5-5} \cmidrule(lr){6-9} \cmidrule(lr){10-12}
        & \textbf{IL} & \textbf{~~~~RL} & \textbf{IL} & \textbf{RL} & CR $\downarrow$ & AF-CR $\downarrow$ & Safety@1 $\uparrow$ & Safety@2 $\uparrow$ & EP-Mean $\uparrow$ & EP@1.0 $\uparrow$ & EP@0.9 $\uparrow$ \\
        \midrule
        1 & \cmark &       &       &       & 0.533 & 0.264 & 0.418 & 0.281 & 0.970 & 0.516 & 0.894 \\
        2 & \cmark & ~~~~\cmark &       &       & \underline{0.287} & \underline{0.104} & \underline{0.682} & \underline{0.582} & 0.955 & 0.391 & 0.824 \\
        3 & \cmark & ~~~~\cmark & \cmark &       & 0.403 & 0.197 & 0.555 & 0.434 & 0.973 & 0.527 & 0.936 \\
        4 & \cmark &       &       & \cmark &  0.337 & 0.166 & 0.615 & 0.496 & \underline{0.987} & \underline{0.728} & \textbf{0.986} \\
        \rowcolor{ourlightblue}
        5 &  \cmark & ~~~~\cmark & \cmark & \cmark & \textbf{0.234} & \textbf{0.092} & \textbf{0.730} & \textbf{0.596} & \textbf{0.988} & \textbf{0.736} & \underline{0.984} \\
        \bottomrule
    \end{tabular}
    \vspace{-4pt}
    \begin{flushleft}
            \footnotesize \textbf{Gen.} = Generator, \textbf{Disc.} = Discriminator.
        \end{flushleft}
\end{table*}

\subsection{Experimental Results}
We evaluate our method across both closed-loop and open-loop benchmarks to assess safety, efficiency, and trajectory quality under diverse driving scenarios and simulation environments. Closed-loop evaluations focus on interactive driving behaviors, while open-loop tests examine the intrinsic accuracy of predicted trajectories. The following sections present detailed results in each setting.

\noindent\textbf{Closed-loop performance in the BEV Warp environment.}
Closed-loop evaluation is critical for assessing the real driving behavior of planning systems. Tab.~\ref{tab:bevwarp-benchmark} reports results in the BEV Warp simulation environment, evaluated on 512 held-out safety-oriented scenarios and 512 efficiency-oriented scenarios. 

In safety-oriented scenarios, our method reduces the collision rate (CR) from 0.533 (ResAD) to 0.234 and the at-fault collision rate (AF-CR) from 0.264 to 0.092, while increasing Safety@1/2 from 0.418/0.281 to 0.730/0.596. For efficiency-oriented scenarios, the Ego Progress Mean (EP-Mean) improves from 0.970 to 0.988, with route completion metrics EP@1.0/0.9 rising from 0.516/0.894 to 0.736/0.984. These results indicate that the proposed framework substantially enhances both safety and efficiency in closed-loop driving.

\noindent\textbf{Closed-loop evaluation in the 3DGS environment.}
Tab.~\ref{tab:3dgs-benchmark} reports results in the photorealistic 3D Gaussian Splatting (3DGS) benchmark, focusing on safety-oriented driving scenarios. Compared with recent systems including Senna, Senna-2, and RAD, our method achieves the lowest collision rate (0.250) and the highest Safety@1 and Safety@2 (0.723 and 0.644, respectively) among the evaluated methods. These results indicate that the trajectory scoring and reinforcement learning framework is also effective in photorealistic simulation environments.

\noindent\textbf{Open-loop trajectory evaluation.}
Tab.~\ref{tab:ol-comparison} presents open-loop comparisons on the Senna-2 evaluation benchmark, which includes six driving scenarios: car-following start, car-following stop, lane changing, intersections, curves, and heavy braking. Our method substantially reduces collision rates, with the overall CR decreasing to 0.142\%, compared to 0.288\% from the previous best (Senna-2). Both dynamic and static collision components are also lower. FDE and ADE are also reduced to 0.553\,m and 0.208\,m, respectively, further supporting the improved trajectory quality in these open-loop scenarios.

\subsection{Scaling Behavior Analysis}
\begin{figure}[h]
\centering
\includegraphics[width=1.0\linewidth]{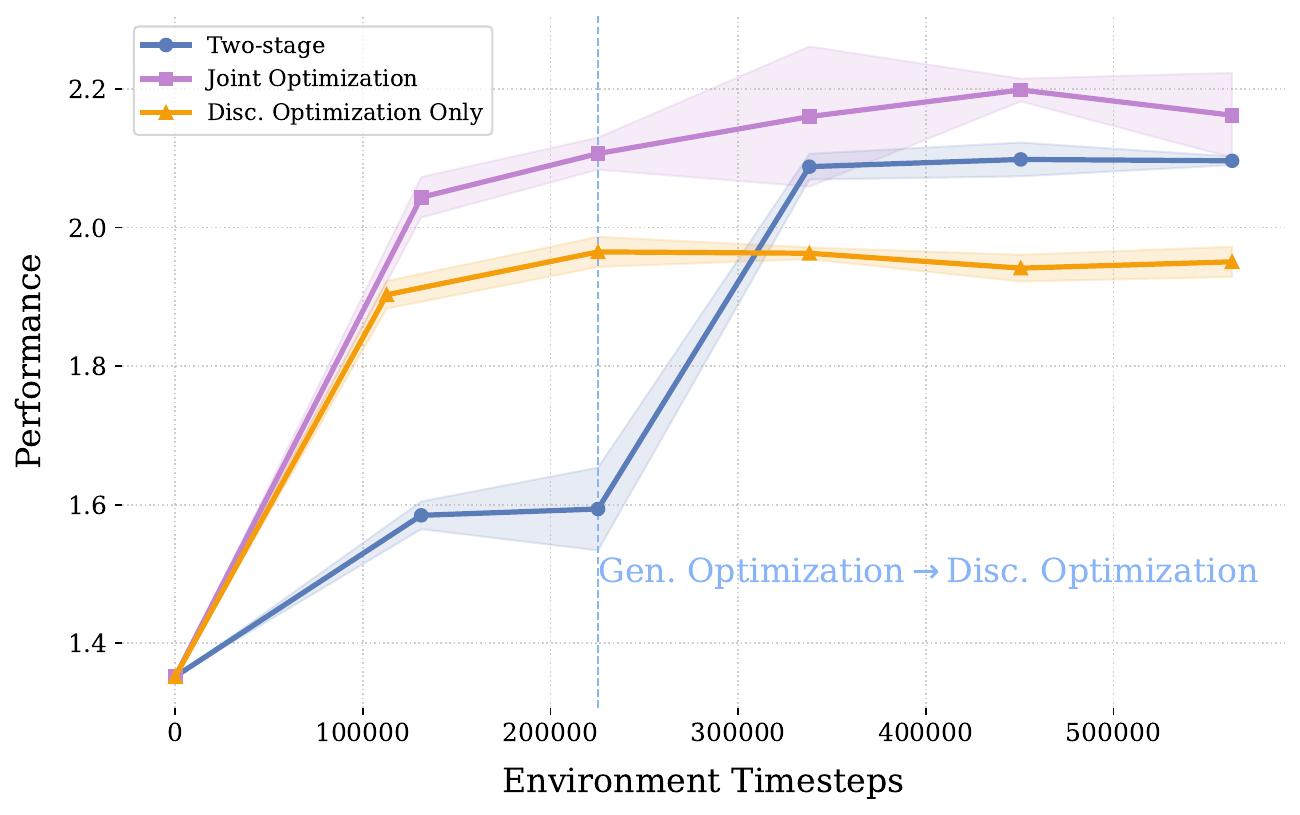} 
    \caption{\textbf{Scaling behavior of training paradigms.} Performance (defined as $2 \times \text{Safety@1} + \text{EP@1.0}$) is evaluated against the cumulative number of environment timesteps. The vertical dashed line marks the transition from on-policy generator optimization to discriminative optimization.}
\label{fig:scale-data}
\end{figure}

To study the scaling behavior of reinforcement learning performance with the cumulative number of environment timesteps, we compare three training strategies: discriminator-only optimization, sequential two-stage training, and synergistic joint optimization. As shown in Fig.~\ref{fig:scale-data}, discriminator-only optimization exhibits limited scaling benefits, as the fixed generator constrains the quality and diversity of rollouts.
The two-stage pipeline improves performance by first optimizing the generator and then training the discriminator, but suffers from suboptimal data utilization due to decoupled updates.
In contrast, joint optimization achieves superior scaling efficiency, yielding a steeper performance curve and better final performance. Under this paradigm, the generator and discriminator are updated in a tightly coupled manner, enabling mutual improvement and progressively enhancing rollout quality. Moreover, both components are trained on shared rollouts, resulting in improved data efficiency and faster convergence.

\subsection{Analysis of RL Training Design}
\noindent\textbf{Ablation on Training Pipeline.} 
We ablate the training pipeline to examine the contributions of generator fine-tuning and discriminator training, as shown in Tab.~\ref{tab:ablation_training_stages}. Incorporating On-policy Generator Optimization improves safety, reducing the collision rate from 0.533 to 0.287, but slightly decreases cruise efficiency. Introducing an IL fine-tuning stage during generator optimization restores cruise performance while maintaining these safety gains. Training the discriminator with RL further enhances both safety and efficiency, lowering the collision rate to 0.337 and increasing EP-Mean to 0.987. By jointly applying RL training to the discriminator and fine-tuning the generator, the proportion of feasible trajectories is increased, allowing discriminator to fully exploit these trajectories and achieve the lowest collision rate (0.234) and highest Safety@1 (0.730) while maintaining high driving efficiency.

\noindent\textbf{Analysis of RL Design Choices.}
We conduct ablations to analyze several design choices across the RL training pipeline, including data collection, sample filtering, model initialization, and optimization objectives.

\begin{table}[h]
    \centering
    \setlength{\tabcolsep}{4pt}
    \caption{\textbf{Sensitivity to execution horizon $H_{\text{reuse}}$.} An execution horizon of 8 provides the optimal balance between stable credit assignment and reactive flexibility for stable RL optimization.}
    \label{tab:trajectory_reuse}
    \begin{tabular}{l|ccc}
        \toprule
        $H_{\text{reuse}}$ & CR $\downarrow$ & Safety@1 $\uparrow$ & EP@1.0 $\uparrow$ \\
        \midrule
        2  & 0.355 & 0.580 & 0.701 \\
        4  & \textbf{0.324} & \textbf{0.627} & 0.604 \\
        \rowcolor{ourlightblue}
        8  & 0.337 & \underline{0.615} & \underline{0.728} \\
        16 & \underline{0.332} & 0.596 & \textbf{0.744} \\
        \bottomrule
    \end{tabular}
\end{table}

\begin{table}[h]
    \centering
    \setlength{\tabcolsep}{4pt}
    \caption{\textbf{Effectiveness of reward-variance-based clip filtering.} By discarding low-variance clips, the model focuses on scenarios with informative training signals, leading to significantly higher efficiency (EP@1.0) without compromising safety.}
    \label{tab:dynamic_sampling_ablation}
    \begin{tabular}{l|ccc}
        \toprule
        Clip Filtering & CR $\downarrow$ & Safety@1 $\uparrow$ & EP@1.0 $\uparrow$ \\
        \midrule
        Disabled & 0.340 & \textbf{0.617} & 0.662 \\
        \rowcolor{ourlightblue}
        Enabled & \textbf{0.337} & 0.615 & \textbf{0.728} \\
        \bottomrule
    \end{tabular}
\end{table}

We evaluate the execution horizon $H_{\text{reuse}}$ in Tab.~\ref{tab:trajectory_reuse}. Lower values (e.g., $2$) increase mode-switching frequency, which weakens the correlation between policy decisions and long-term outcomes, leading to unstable gradients. Conversely, a larger horizon ($16$) enhances temporal consistency but may limit reactivity. An execution horizon of $8$ provides an optimal balance between credit assignment and reactive flexibility, achieving the best performance.

\begin{figure}[ht]
\centering
\vspace{1mm}
\includegraphics[width=1.0\linewidth]{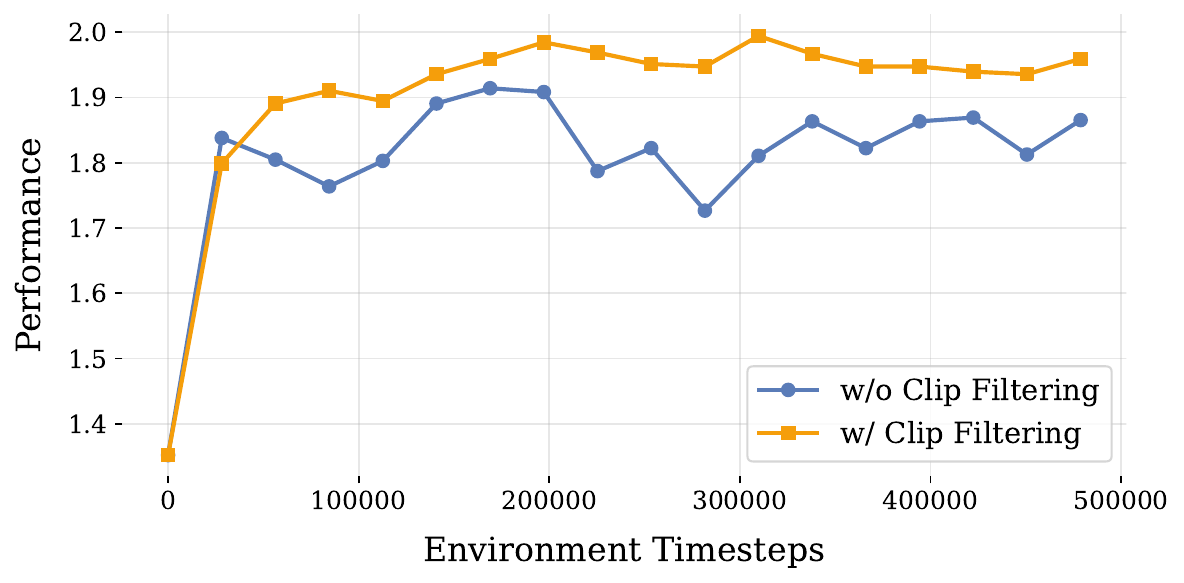} 
\caption{\textbf{Training stability with and without clip filtering.} Compared to the baseline (blue), clip filtering (orange) significantly stabilizes training dynamics and improves performance by prioritizing more informative training signals.}
\label{fig:training_stability}
\end{figure}

Next, we evaluate the impact of reward-based clip filtering for training data selection. As previously detailed, scenarios exhibiting low reward variance across rollouts are discarded, as they offer limited exploratory value. Tab.~\ref{tab:dynamic_sampling_ablation} demonstrates that this filtering mechanism enhances efficiency without compromising safety. Furthermore, as illustrated by the training dynamics in Fig.~\ref{fig:training_stability}, the exclusion of non-informative clips fosters more stable convergence. This confirms that prioritizing scenarios with high-variance outcomes provides denser optimization signals.

\begin{table}[h]
    \centering
    \setlength{\tabcolsep}{4pt}
    \caption{\textbf{Ablation on discriminator model initialization strategies.} Initializing the discriminator from the pre-trained planning head provides a robust structural prior, which is crucial for achieving high safety scores compared to random initialization.}
    \label{tab:scorer_init_ablation}
    \resizebox{\linewidth}{!}{
    \begin{tabular}{l|ccc}
        \toprule
        Discriminator Initialization & CR $\downarrow$ & Safety@1 $\uparrow$ & EP@1.0 $\uparrow$ \\
        \midrule
        Random & 0.426 & 0.512 & 0.710 \\
        \rowcolor{ourlightblue}
        From Planning Head & \textbf{0.337} & \textbf{0.615} & \textbf{0.728} \\
        \bottomrule
    \end{tabular}
    }
\normalsize
\end{table}

\begin{table}[ht]
    \centering
    \setlength{\tabcolsep}{4pt}
    \caption{\textbf{Effect of group size on TC-GRPO optimization.} A group size of 4 achieves the best balance between reward normalization and safety performance.}
    \label{tab:rollouts_per_group}
    \normalsize
    \begin{tabular}{l|ccc}
        \toprule
        Group Size & CR $\downarrow$ & Safety@1 $\uparrow$ & EP@1.0 $\uparrow$ \\
        \midrule
        2  & 0.313 & 0.639 & 0.729 \\
        \rowcolor{ourlightblue}
          4  & \textbf{0.234} & \textbf{0.730} & \underline{0.736} \\
        8  & \underline{0.291} & \underline{0.676} & \textbf{0.756} \\
        \bottomrule
    \end{tabular}
\end{table}

We further investigate the impact of discriminator initialization on optimization stability. As shown in Tab.~\ref{tab:scorer_init_ablation}, initializing the discriminator from the planning head significantly improves both safety and efficiency, whereas random initialization results in a drop in safety and  a slight degradation in efficiency, indicating imbalanced optimization behavior. This performance gap stems from the structural prior provided by the pre-trained weights, enabling more reliable scene understanding and trajectory evaluation. 

In addition, we evaluate the number of rollouts per group within the TC-GRPO in Tab.~\ref{tab:rollouts_per_group}. Under our current setup, a group size of $4$ yields the best performance, specifically achieving the lowest collision rate and highest Safety@$1$. While increasing the group size to $8$ marginally improves efficiency (EP@$1.0$), it results in a performance drop across safety metrics. Consequently, we adopt a group size of $4$ to ensure a robust balance between safety and efficiency.

\begin{figure}[ht]
\centering
\vspace{1mm}
\includegraphics[width=1.0\linewidth]{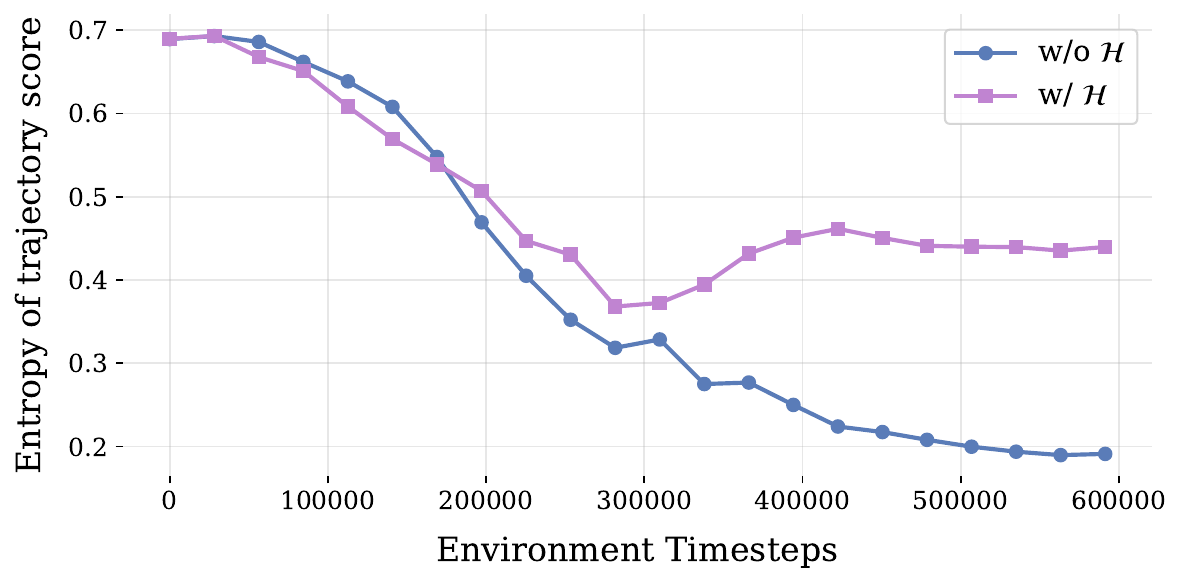} 
\caption{\textbf{Entropy stability analysis.} The inclusion of $\mathcal{H}$ (purple) prevents a rapid decline in trajectory score entropy, maintaining a more diverse distribution for stable optimization compared to the baseline without regularization (blue). }
\label{fig:rl_entropy_stability}
\end{figure}

Finally, we examine the role of the entropy term $\mathcal{H}$ in the RL objective.
Including $\mathcal{H}$ prevents trajectory scores from collapsing to extreme values near 0 or 1, producing a more balanced distribution that facilitates stable RL optimization, as illustrated in Fig.~\ref{fig:rl_entropy_stability}. 
This improved stability translates into lower collision rates and higher Safety@1 and EP@1.0, as reported in Tab.~\ref{tab:rl_entory}.

\begin{table}[h]
    \centering
    \setlength{\tabcolsep}{4pt}
    \caption{\textbf{Ablation on entropy regularization.} The entropy term $\mathcal{H}$ in RL objective maintains a diverse scoring distribution and prevents policy collapse. }
    \label{tab:rl_entory}
    \begin{tabular}{l|ccc}
        \toprule
        RL Objective & CR $\downarrow$ & Safety@1 $\uparrow$ & EP@1.0  $\uparrow$ \\
        \midrule
        $\mathcal{J}_{\text{RL}}$ (w/o $\mathcal{H}$) & 0.254 & 0.697 & 0.727 \\
        \rowcolor{ourlightblue}
        $\mathcal{J}_{\text{RL}}$ (w/ $\mathcal{H}$)    & \textbf{0.234} & \textbf{0.730} & \textbf{0.736} \\
        \bottomrule
    \end{tabular}
\end{table}

\begin{figure}[ht]
\centering
\vspace{1mm}
\includegraphics[width=1.0\linewidth]{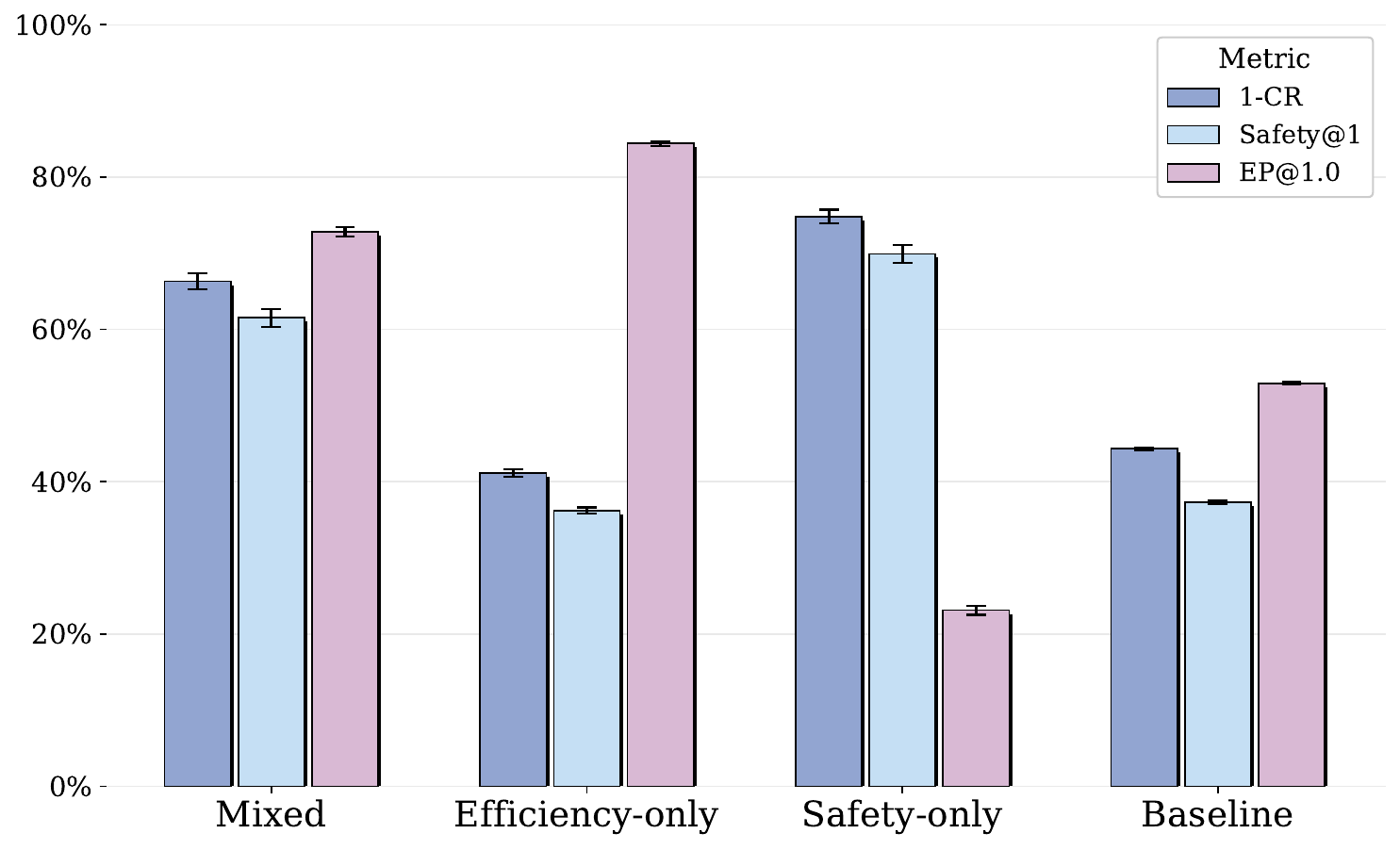} 
\caption{\textbf{Impact of scenario composition on closed-loop performance.} Compared to single-objective training (\textit{Safety-only} or \textit{Efficiency-only}), the \textit{Mixed} strategy achieves a superior trade-off across all metrics. Models trained on biased subsets exhibit significant performance drops in complementary tasks (e.g., \textit{Efficiency-only} lacks safety robustness), highlighting the necessity of diverse scenario curation for balanced driving policies.}
\label{fig:data_odo_ablation}
\end{figure}

\noindent\textbf{Ablation on Training Scenario Composition.} 
We investigate the impact of training data distribution by comparing models trained on \textit{Mixed} scenarios (integrating safety- and efficiency-oriented clips) against those trained on single-objective subsets (\textit{Safety-only} or \textit{Efficiency-only}) and a \textit{Baseline} configuration (Fig.~\ref{fig:data_odo_ablation}). 
Empirical results show that the \textit{Mixed} configuration achieves the most balanced performance, effectively navigating the trade-off between collision avoidance and navigation progress. 
In contrast, training exclusively on \textit{Efficiency-oriented} data maximizes the imitation accuracy (EP@1.0) but suffers from substantially degraded safety robustness, indicating a failure to handle high-risk interactions. Conversely, \textit{Safety-oriented} training prioritizes risk mitigation at the expense of efficiency. These findings suggest that single-objective training induces a significant performance bias, whereas incorporating a diverse scenario composition enables the model to learn a more robust driving policy that generalizes across competing objectives.

\noindent\textbf{Inference-time Scaling Analysis.}
We investigate the computational scalability of \thename{} by varying the candidate trajectory count $M$ at inference (Tab.~\ref{tab:candidate_trajectories}). While the model is trained with $M=32$, increasing $M$ consistently scales the navigation efficiency (EP@1.0) from 0.667 to 0.814. Although safety metrics exhibit minor fluctuations due to the increased complexity of the expanded search space, the stable Collision Rate (CR) at higher $M$ values demonstrates a robust inference-time scaling effect. This confirms that the discriminator can effectively leverage additional test-time computation to identify higher-quality trajectories without further retraining.

\begin{table}[ht]
    \centering
    \setlength{\tabcolsep}{4pt}
    \caption{\textbf{Inference-time scaling with candidate count $M$.} Expanding the candidate pool during inference consistently improves planning performance by enabling a more comprehensive search of the multimodal action space.}
    \label{tab:candidate_trajectories}
    \normalsize
    \begin{tabular}{l|ccc}
        \toprule
        $M$ & CR $\downarrow$ & Safety@1 $\uparrow$ & EP@1.0 $\uparrow$ \\
        \midrule
        8  & 0.275 & 0.693 & 0.667 \\
        16  & 0.266 & 0.689 & 0.711 \\
        \rowcolor{ourlightblue}
        32  & \textbf{0.234} & \textbf{0.730} & 0.736 \\
        64 & \underline{0.252} & 0.699 & \textbf{0.816} \\
        128 & \textbf{0.234} & \underline{0.719} & \underline{0.814} \\
        \bottomrule
    \end{tabular}
\end{table}

\begin{figure*}[ht]
\centering
\includegraphics[width=1.0\textwidth]{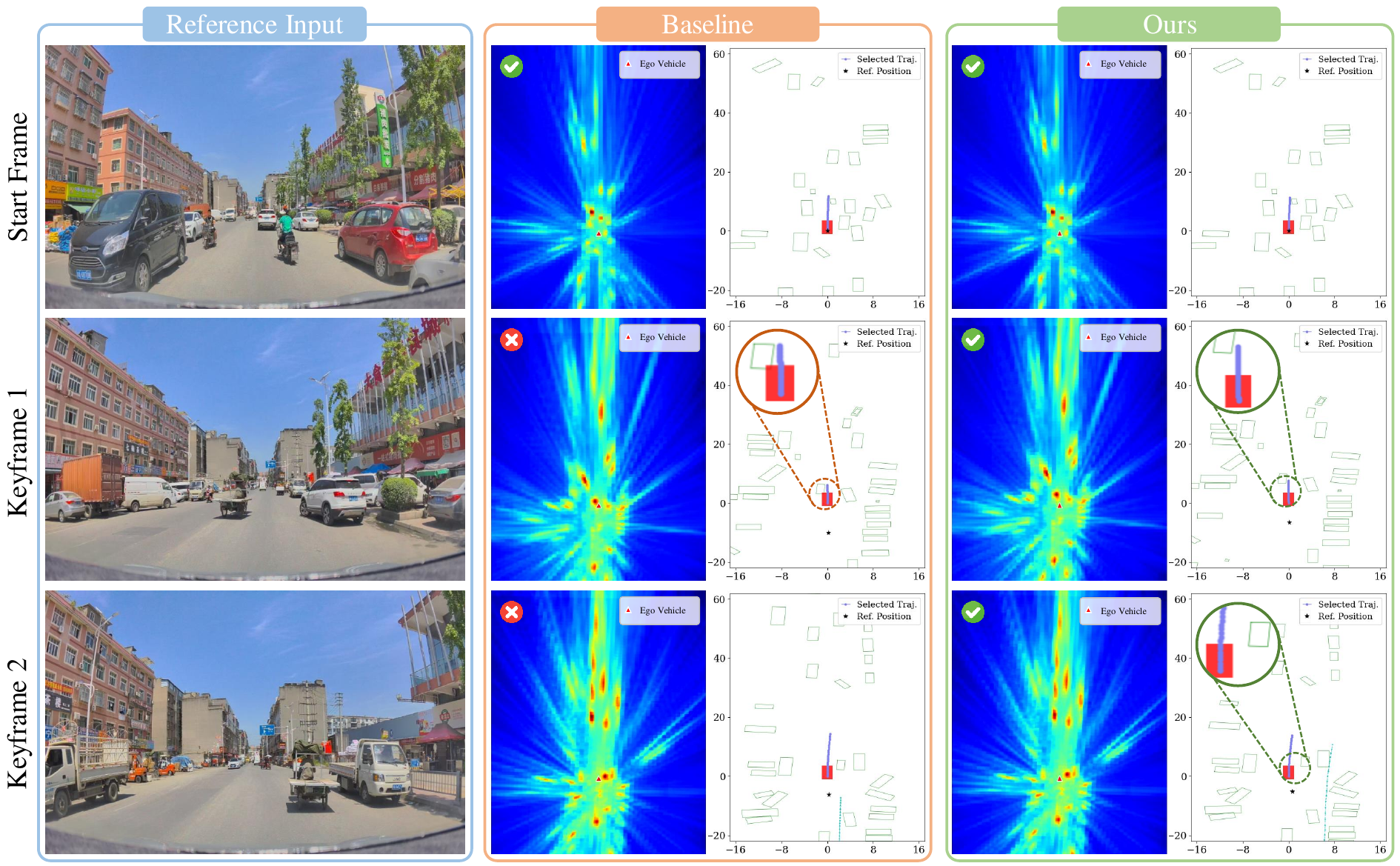} 
\caption{\textbf{Qualitative comparison of closed-loop safety interaction.} (1) \textit{Reference Input} displays the front-view camera feed. (2) \textit{Baseline} and (3) \textit{Ours} each show the warped BEV input (left) and the corresponding perception and planning output (right). Starting from the same initial state at the \textit{Start Frame}, the two models exhibit diverging behaviors as the interaction unfolds. By \textit{Keyframe 1}, the \textit{Baseline} fails to mitigate the risk and results in a collision, while our model performs proactive deceleration to maintain safety. At \textit{Keyframe 2}, our vehicle resumes stable navigation after the threat clears, validating the robustness of our learned policy.}
\label{fig:safety}
\end{figure*}

\subsection{Qualitative Comparisons.}
\begin{figure*}[ht]
\centering
\includegraphics[width=1.0\textwidth]{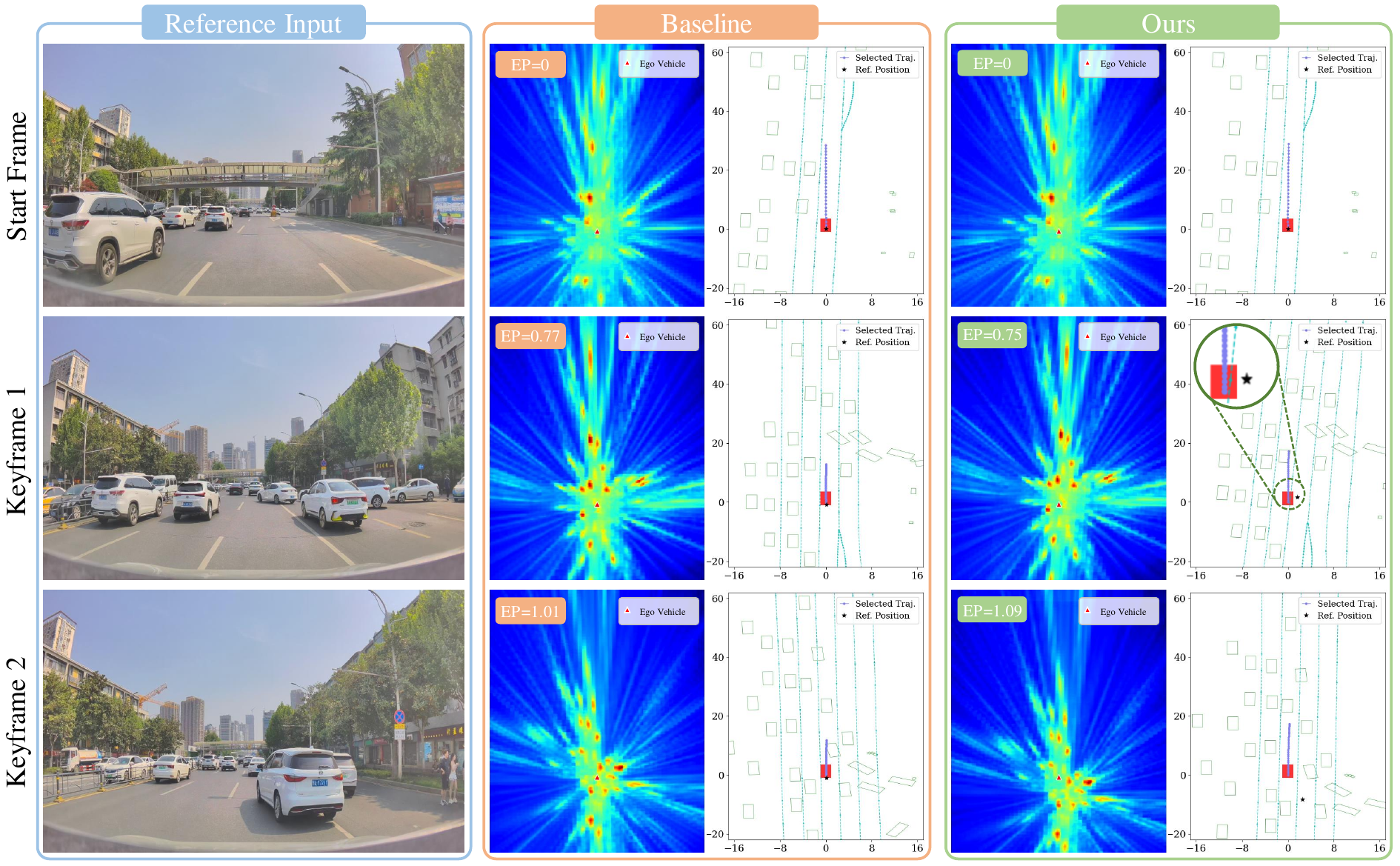} 
\caption{\textbf{Qualitative comparison of driving efficiency in dynamic traffic.} (1) \textit{Reference Input} displays the front-view camera feed. (2) \textit{Baseline} and (3) \textit{Ours} each show the warped BEV input (left) and the perception and planning output with Efficiency Progress (EP) (right). Starting from the same initial state, both models encounter a vehicle merging from the right at \textit{Keyframe 1}. Despite sufficient clearance in the adjacent lane, the \textit{Baseline} exhibits overly conservative behavior, decelerating to wait behind the merging vehicle. In contrast, our model executes a proactive lane change to bypass the slow-moving agent. By \textit{Keyframe 2}, our model achieves superior navigational progress ($\text{EP}=1.09$) compared to the \textit{Baseline} ($\text{EP}=1.01$), validating its ability to make efficient tactical decisions in interactive environments.}
\label{fig:efficiency}
\end{figure*}
We provide qualitative comparisons of closed-loop performance in Fig.~\ref{fig:safety} and Fig.~\ref{fig:efficiency}. As shown in Fig.~\ref{fig:safety}, RAD-2 exhibits superior safety in critical interactions by executing proactive deceleration to avoid collisions that the baseline fails to mitigate. Furthermore, Fig.~\ref{fig:efficiency} demonstrates that our model achieves higher driving efficiency in dynamic traffic through agile lane-changing maneuvers. These results validate the effectiveness of the proposed framework in balancing safety and progress within interactive environments.

\section{Limitations and Future Work}
Despite the substantial improvements in closed-loop stability and safety, several limitations of the proposed framework merit further discussion and provide avenues for future research.

\noindent\textbf{Representation Specificity.} 
The efficiency of our BEV-Warp simulation environment is fundamentally rooted in the manipulation of BEV feature maps. While this design facilitates high-throughput policy iteration for systems that explicitly rely on BEV-centric perception, its applicability is constrained for architectures that utilize raw camera pixels or unified latent embeddings without explicit spatial-equivariant grid structures. In such cases, the geometric warping mechanism would require a more generalized transformation module or a direct latent-space world model.

\noindent\textbf{Transition to Generative World Models.} 
A promising extension of this work is the integration of our optimization pipeline with Generative World Models (WM). Although current WM offer superior flexibility and photorealism compared to feature-level warping, they often suffer from significant computational overhead and temporal drift during long-horizon generation, which limits their utility for large-scale RL training. Future efforts will focus on optimizing the inference efficiency and temporal consistency of latent-based world models. By porting our framework into a more flexible generative simulator, we aim to further scale the diversity of training scenarios and bridge the remaining fidelity gap between simulation and the complex, open-world dynamics of real-world driving.

\section{Conclusion}
In this work, we presented \thename{}, a unified generator–discriminator framework for stable reinforcement learning in diffusion-based motion planning. We introduced Temporally Consistent Group Relative Policy Optimization to improve credit assignment through temporally coherent sampling, and On-policy Generator Optimization to progressively refine trajectory distributions using structured feedback. To support efficient large-scale training, we further introduce BEV-Warp, a feature-level simulation pipeline enabling scalable closed-loop learning. Extensive experiments demonstrate that \thename{} consistently improves both safety and efficiency across diverse benchmarks, achieving substantial reductions in collision rates while maintaining reliable closed-loop navigation performance. Ablation studies further verify the effectiveness of the proposed generator–discriminator formulation and temporally consistent optimization strategy.

\section{Acknowledgement} 
We would like to acknowledge Hui Sun, Zhihao Guan, Songlin Yang, Qingjie Wang, Zhengqing Chen, Xiaoyang Guo, Xinbang Zhang and Nuoya Zhou for valuable discussions and assistance, Xinhui Bai, Wei Li and Pipi Ke for real-world closed-loop evaluation, and Zehua Li and Cheng Chi for  data infra support.

{
    \small
    \bibliographystyle{ieeenat_fullname}
    \bibliography{main}
}


\end{document}